\documentclass[lettersize,journal]{IEEEtran}
\usepackage[T1]{fontenc}
\usepackage[english]{babel}
\usepackage{times} 
\usepackage{indentfirst}
\usepackage{verbatim}
\usepackage{blindtext}
\usepackage{balance} 
\usepackage[a4paper, top=2.5cm, bottom=2cm, left=1.7cm, right=1.7cm, marginparwidth=1.75cm]{geometry}

\usepackage{amsmath, amsfonts}
\usepackage{algorithmic}
\usepackage{algorithm}
\usepackage{textcomp} 
\usepackage{import}
\usepackage{standalone}

\usepackage{array}
\usepackage{booktabs}
\usepackage{multirow}
\usepackage{makecell}
\usepackage{tabularx}
\usepackage{tabu}
\usepackage{adjustbox}
\usepackage{ragged2e}
\usepackage[flushleft]{threeparttable}
\newcolumntype{Y}{>{\raggedright\arraybackslash}X}
\newcolumntype{P}[1]{>{\color{\currentcolor}}p{#1}}

\usepackage{graphicx}
\usepackage[table]{xcolor}
\usepackage{float}
\usepackage{stfloats}
\usepackage{placeins} 
\usepackage[caption=false, font=normalsize, labelfont=sf, textfont=sf]{subfig}
\usepackage{tikz}
\usetikzlibrary{calc, positioning}

\usepackage[most]{tcolorbox}
\usepackage{mdframed}
\usepackage{listings}
\usepackage{soul}
\usepackage[normalem]{ulem}
\usepackage{enumitem}

\usepackage{cite}
\usepackage{url}

\soulregister{\cite}{1}

\mdfdefinestyle{RefBoxStyle}{
    linecolor=black,
    linewidth=1pt,
    roundcorner=5pt,
    innertopmargin=10pt,
    innerbottommargin=10pt,
    innerleftmargin=10pt,
    innerrightmargin=10pt,
    backgroundcolor=gray!10,
    everyline=true,
    userdefinedwidth=\textwidth
}

\renewmdenv[
  leftmargin=0.4em,
  rightmargin=0.4em,
  innerleftmargin=0.6em,
  innerrightmargin=0.6em
]{quote}

\begin{document}

\title{Spiking Neural Network Architecture Search:\\A Survey}

\author{Kama Svoboda and Tosiron Adegbija,~\IEEEmembership{Senior Member,~IEEE} 

\thanks{The authors are with the Department of Electrical and Computer Engineering, The University of Arizona, USA, email: \{ksvoboda, tosiron\}@arizona.edu.}
}



\maketitle
\begin{abstract}
This survey paper presents a comprehensive examination of Spiking Neural Network (SNN) architecture search (SNNaS) from a unique hardware/software co-design perspective. SNNs, inspired by biological neurons, have emerged as a promising approach to neuromorphic computing. They offer significant advantages in terms of power efficiency and real-time resource-constrained processing, making them ideal for edge computing and IoT applications. However, designing optimal SNN architectures poses significant challenges, due to their inherent complexity (e.g., with respect to training) and the interplay between hardware constraints and SNN models. We begin by providing an overview of SNNs, emphasizing their operational principles and key distinctions from traditional artificial neural networks (ANNs). We then provide a brief overview of the state of the art in NAS for ANNs, highlighting the challenges of directly applying these approaches to SNNs. We then survey the state of the art in SNN-specific NAS approaches. Finally, we conclude with insights into future research directions for SNN research, emphasizing the potential of hardware/software co-design in unlocking the full capabilities of SNNs. This survey aims to serve as a valuable resource for researchers and practitioners in the field, offering a holistic view of SNNaS and underscoring the importance of a co-design approach to harness the true potential of neuromorphic computing.

\end{abstract}


\section{Introduction}
As the demand for artificial intelligence (AI) in resource-constrained environments grows, the need for energy-efficient neural architectures has become increasingly critical. Spiking neural networks (SNNs), inspired by the brain's biological processes, have emerged as a promising solution for low-power AI applications, particularly in edge computing and real-time processing scenarios. These networks, which communicate through discrete temporal spikes rather than continuous values, represent a fundamental departure from conventional artificial neural networks (ANNs) and align closely with the principles of neuromorphic computing, an area that has attracted significant research attention for its potential to enable ultra-low-power AI systems.

The architecture of an SNN plays a critical role in both its performance and efficiency \cite{na2022autosnn}. However, designing optimal architectures poses unique challenges due to the intricate interplay between network topology, neuronal dynamics, and hardware constraints. Unlike traditional ANNs, SNNs employ discrete spike-based communication, necessitating specialized design and training strategies \cite{kim2022neural, tavanaei2019deep}. Most SNN architectures have followed one of two suboptimal paths: either adapting existing ANN architectures (such as VGG-Net~\cite{simonyan2015deep} or ResNet~\cite{he2016deep}) that were not designed for spike-based communication, or relying on expert-driven manual design through time-consuming trial-and-error processes \cite{kim2022neural, na2022autosnn, putra2025spikenas}. Neither approach fully capitalizes on the unique characteristics of spiking neurons.

\begin{figure*}[t!]
		\centering
		\includegraphics[width=0.8\linewidth]{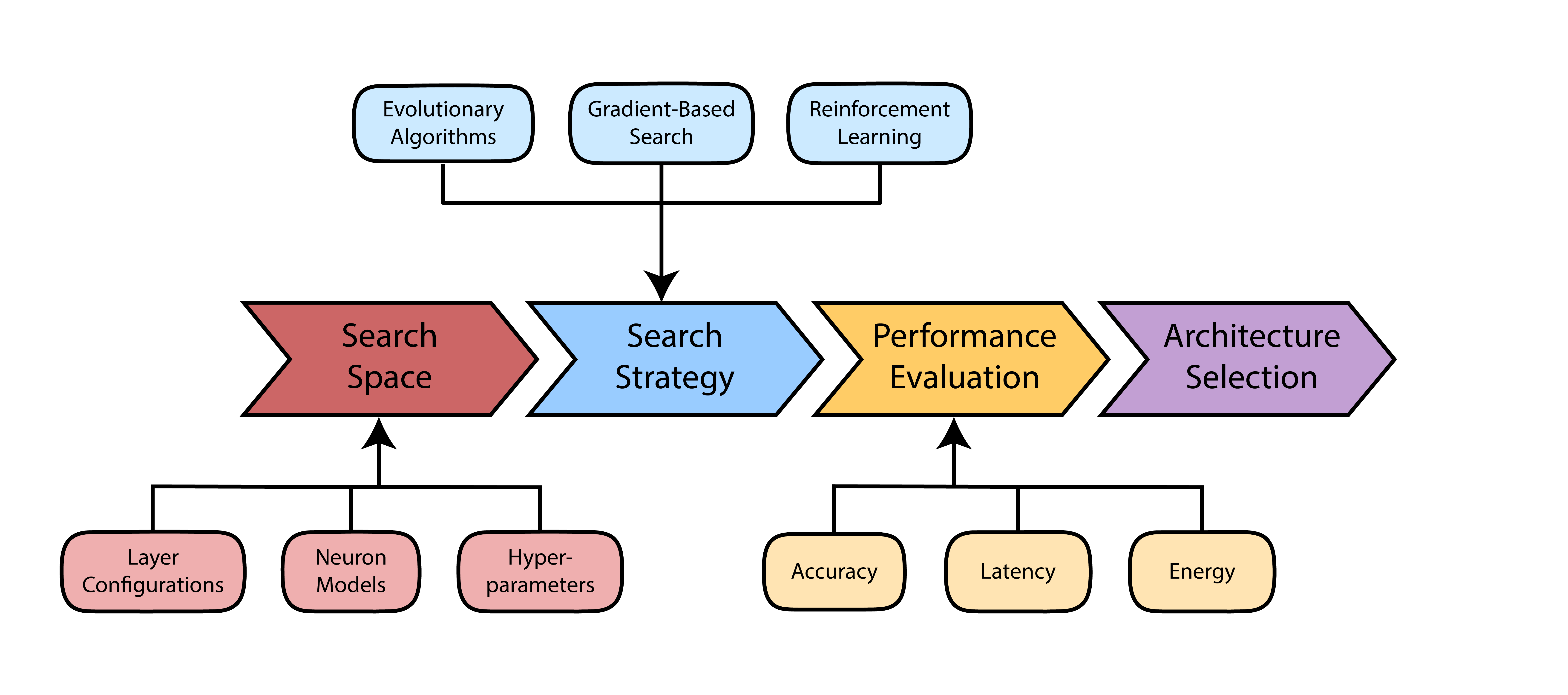}
		\vspace{-20pt}
		\caption{
        Overview of the NAS workflow for SNNs. The pipeline defines the three NAS components within an SNN-specific roadmap: 1) search space, 2) search strategy, and 3) evaluation \cite{elsken2019neural}. The search space defines architectural choices such as layer configurations, neuron models, and hyperparameters; search strategies explore this space (e.g., evolutionary algorithms, gradient-based methods, and reinforcement learning); and candidate architectures are evaluated using performance metrics including accuracy, latency, and energy to guide final architecture selection.
        }	\label{fig:roadmap}
\end{figure*}

Given these challenges, \textit{Spiking Neural Network Architecture Search (SNNaS)} has emerged as a critical research direction \cite{masters, na2022autosnn, che2022differentiable, gaq, kim2022neural, che2023auto, shen2023brain, xie2023efficient, liu2024unleashing, putra2025spikenas, putra2024methodology, putra2024neuronas, pan2024brain, yan2024efficient, liu2024lite, padovano2024spikeexplorer, ajay2024despine, wang2024autost}. As illustrated in Figure \ref{fig:roadmap}, SNNaS methods comprise three core components: a search space defining architectural choices (e.g., layer configurations, neuron models, hyperparameters), a search strategy for exploring this space (e.g., evolutionary algorithms, gradient-based methods, reinforcement learning), and a performance evaluation mechanism that guides architecture selection based on metrics such as accuracy, latency, and energy consumption. SNNaS methods focus on automating the discovery of optimal SNN architectures by simultaneously optimizing multiple, often competing objectives, including inference accuracy, processing speed, and energy consumption. \textit{Importantly, this optimization process must consider both software and hardware constraints, particularly when targeting deployment on neuromorphic hardware} \cite{teich2012hardware}.

Despite the growing body of literature on spiking neural networks, most existing surveys focus on foundational aspects such as neuron models, training methodologies, learning rules, hardware implementations, or application domains. These works provide essential background on how SNNs operate and how they can be trained or deployed, but the problem of automating architectural design choices---long recognized as critical for ANN performance and efficiency---has only recently begun to receive focused attention in the SNN domain.

This survey paper provides a comprehensive analysis of SNNaS research through the lens of hardware-software co-design. We begin by establishing fundamental SNN concepts, including spike encoding/decoding mechanisms, neuron models, and training methodologies, while highlighting key distinctions from traditional ANNs. Building on this foundation, we examine how Neural Architecture Search (NAS) approaches from the ANN domain must be adapted for the unique characteristics of SNNs. The survey's core contribution lies in its systematic exploration of SNNaS methodologies, encompassing search space design, search algorithms, acceleration strategies, and performance estimation techniques, with particular emphasis on integrating hardware constraints. Finally, looking forward, we identify promising research directions, including surrogate gradient optimization, Bayesian approaches, hybrid methodologies, and emerging neuromorphic hardware platforms. Through this analysis, we aim to illuminate both the challenges and opportunities in developing efficient SNN architectures for next-generation AI systems.

In summary, the key insights of this survey are as follows:

\begin{enumerate}
    \item \textbf{Hardware-software co-design is essential, not optional.} SNN-specific temporal dynamics and event-driven computation create fundamental mismatches with ANN-centric NAS methods, requiring integrated consideration of algorithmic and hardware constraints from the outset.
    \item \textbf{Search space design critically determines SNNaS success.} The field has evolved from naively replacing ReLu with LIF neurons toward SNN-specific representations (e.g., cell-based directed acyclic graphs DAGs, hierarchical structures, and biologically-inspired motifs) that better capture spiking computation.
    \item \textbf{Temporal dynamics remain underexploited.} Despite being a defining advantage of SNNs, many of the surveyed SNNaS methods adapt ANN-centric approaches that overlook time-related factors such as spike timing, timestep optimization, and temporal attention.
    \item \textbf{Training-free evaluation strategies are transforming efficiency.} Zero-shot metrics, surrogate models, and one-shot supernets have reduced search costs by orders of magnitude, making SNNaS practical for resource-constrained settings.
    \item  \textbf{Multi-objective optimization is inherent to SNNaS.} Unlike traditional NAS, SNNaS must simultaneously balance accuracy, energy, latency, memory, and spike sparsity. This requirement aligns naturally with evolutionary and Bayesian approaches.
    \item \textbf{Co-exploration outperforms sequential optimization.} Frameworks that jointly search neural architectures and hardware configurations consistently achieve better energy-delay products than methods optimizing these dimensions separately.
\end{enumerate}

\section{Background} \label{sec:background}

Spiking Neural Networks (SNNs) mimic biological nervous systems, using spiking neurons to transmit data through precisely timed electrical pulses \cite{gerstner2002spiking}. This temporal element provides fresh perspectives on brain dynamics and neural network efficiency. SNNs excel at processing time-dependent patterns, making them ideal for complex recognition tasks \cite{Ghosh-Dastidar2009Spiking}. By emulating nature's design, these networks offer a powerful approach to artificial intelligence that leverages the intricacies of biological information processing.

\subsection{Key Components Affecting Architecture Search}
\subsubsection{Neuron Models}
SNNs employ diverse neuron models that trade biological plausibility for computational efficiency. The Leaky Integrate-and-Fire (LIF) model~\cite{koch1998methods, gerstner2002spiking} strikes a practical balance between these factors and therefore dominates SNNaS research. In contrast, the Hodgkin–Huxley model~\cite{hodgkin1952quantitative} offers high biophysical fidelity but incurs substantial computational cost. The Izhikevich model~\cite{izhikevich2003simple} occupies an intermediate position, reproducing a wide range of spiking behaviors with relatively low computational overhead~\cite{taherkhani2020review}, albeit without detailed biophysical interpretability. Crucially, this trade-off between fidelity and simplicity propagates beyond modeling, directly affecting architecture search space complexity and hardware implementation efficiency.

\subsubsection{Spike Encoding} Information encoding significantly impacts architecture performance. SNNs encode information using discrete neuronal events or spike trains, in which precise spike patterns and timing are used to represent data. 

Various encoding algorithms transform input data into spike trains, each suited to specific applications and hardware constraints \cite{almomani2019comparative, aliyev2024_sparsityOverview}. Rate-based encoding, for example, uses the firing frequency to represent signal magnitude \cite{guo2021neural}. Temporal encoding methods focus on the timing of spikes, offering different strategies: time-to-first-spike emits earlier spikes for larger signals \cite{park2020t2fsnn}; phase coding periodically modulates spike weights \cite{kim2018deep}; burst encoding utilizes inter-spike intervals and bursts to convey information \cite{izhikevich2003bursts}; and spatio-temporal encoding captures both spike timing and spatial distribution across neurons or layers~\cite{sheik2017spike}. 

The choice of encoding scheme significantly influences the efficiency of information transfer and the performance of both hardware and software. Spatio-temporal encoding offers high efficiency for tasks requiring dynamic interaction between neurons \cite{sheik2017spike}, while temporal pattern coding can reduce power consumption in computationally demanding tasks \cite{rueckauer2021temporal}. These trade-offs highlight the importance of aligning encoding strategies with the specific requirements of both the application and the underlying hardware.

\subsubsection{Training Approaches} 
Synaptic plasticity underpins learning in biological brains, with connection strengths changing over time \cite{abbott2000synaptic}. SNNs aim to mimic this but face unique training challenges. The non-differentiable spike function complicates traditional backpropagation methods \cite{shrestha2019review}. Furthermore, training deep SNNs with multi-layer learning is particularly challenging, often resulting in learning being confined to a single layer in many SNNs \cite{masquelier2007unsupervised, beyeler2013categorization, tavanaei2016acquisition}.

Four primary methods used for training SNNs influence architecture design: \textit{surrogate gradients}, \textit{ANN-to-SNN conversion}, \textit{spike-timing-dependent plasticity (STDP)}, and \textit{evolutionary optimization (EO)}.

Surrogate gradients enable SNN training by approximating non-differentiable spike functions, allowing backpropagation while preserving temporal dynamics \cite{neftci2019surrogate}. While backpropagation through time (BPTT) methods have demonstrated high accuracy with low latency on image classification tasks~\cite{neftci2019surrogate, bohte2000spikeprop, wu2018spatio}, their resource intensity limits application to simpler architectures \cite{rathi2021diet}. The optimal choice of surrogate function must balance accuracy and implementation efficiency. Although software implementations offer flexibility \cite{neftci2019surrogate}, hardware considerations impact spiking activity sparsity and implementation efficiency \cite{aliyev2024fine}.

ANN-to-SNN Conversion transforms ANNs into SNNs by training a `shadow' ANN, then replacing analog neurons with spiking neurons, retaining ANN weights, and reconfiguring neurons to mimic the original network. This bypasses the issue of non-differentiable spiking activity and reduces training overhead \cite{wang2022signed}. The spike rate or latency codes approximate the ANN's behavior, but this conversion can require lengthy simulations that can negate SNNs' power and speed advantages~\cite{rueckauer2017conversion}. Moreover, the approximation process may degrade performance, especially as latency decreases \cite{deng2021optimal, ho2021conversion}. Additionally, since ANNs are often trained on static datasets like MNIST and CIFAR-10, the converted SNNs may underutilize temporal dynamics, limiting their potential advantages \cite{deng2020rethinking}. As a result, ANN-to-SNN conversion methods primarily constrain architecture design rather than discover SNN-native structures, and are therefore best interpreted as baselines or transitional approaches rather than fully spiking-native SNNaS methods 

Spike-Timing Dependent Plasticity (STDP)~\cite{bi1998synaptic} adjusts synapse strength based on spike timing as an unsupervised learning rule. While naturally adapted to temporal patterns, its effectiveness diminishes in deeper networks~\cite{vigneron2020critical}, and STDP-trained networks often underperform other algorithms~\cite{liu2021sstdp}.

Evolutionary Optimization for Neuromorphic Systems (EONS)~\cite{schuman2020evolutionary} uses evolutionary optimization (EO), a gradient-free method, to jointly evolve network topology and neuron/synapse parameters (e.g., weights, thresholds, synaptic delays). Because EO does not rely on gradient information, it bypasses the challenges posed by non-differentiable spike activation functions. As a highly versatile approach, it enables SNN design for classification, control, and reservoir generation under neuromorphic hardware constraints~\cite{schuman2016evolutionary}. The disadvantage is that EO methods are often computationally expensive, requiring substantial compute to reach convergence.

\subsection{Challenges of SNNs}
Table \ref{table:SNNs-Advantages-Challenges} illustrates the fundamental paradox of SNNs: while offering compelling advantages like energy efficiency and temporal processing, each benefit introduces corresponding challenges that must be carefully balanced in the design process. First, as previously alluded to, their non-differentiable spiking mechanism complicates the training process, making standard techniques like backpropagation unsuitable without adaptations \cite{deng2020rethinking}. While techniques like surrogate gradients have become a workhorse for training SNNs, they remain less mature and efficient compared to those used in traditional ANNs. Furthermore, many traditional training approaches fail to account for the unique restrictions of neuromorphic hardware, such as physical connectivity limitations, limited synaptic weight resolution, and the presence of operational noise. These discrepancies can lead to significant performance degradation when a trained network is eventually mapped onto a physical system.

SNNs can be computationally costly due to their temporal dynamics. Unlike ANNs, which compute activations in a single pass, SNNs require multiple time steps to capture spike-based neuronal dynamics, increasing both computational cost and latency \cite{valadez2017step}. This complexity is further exacerbated during the training phase. For instance, evolutionary approaches like EONS~\cite{schuman2016evolutionary} can require a large number of epochs and a large population size to converge, necessitating substantial compute resources and parallelization. Additionally, a "hardware/software boundary" bottleneck may arise if a software simulator is unavailable, due to the high communication overhead required to configure and evaluate diverse network structures on neuromorphic hardware \cite{schuman2020evolutionary}. However, these challenges can be partially mitigated through multi-objective optimization, which allows for the evolution of non-traditional, highly simplified network structures that use much fewer neurons and synapses than standard architectures, thereby improving area and power efficiency \cite{schuman2020evolutionary}.

Additionally, SNN research suffers from a lack of standardized tools, models, and best practices. Unlike ANNs, which benefit from mature ecosystems such as TensorFlow and PyTorch, SNNs rely on emerging tools like NEST, BindsNET, Brian, and SNNtorch, which are still evolving \cite{cramer2020heidelberg, pfeiffer2018deep}. This disparity hinders widespread adoption and slows down the development of optimized solutions. Similarly, SNNs often fail to match ANN-level accuracies on traditional benchmarks like ImageNet and CIFAR, due to the nature of these datasets, which are acquired using frame-based sensors rather than event-driven cameras \cite{pfeiffer2018deep, nunes2022spiking}. This discrepancy highlights the need for benchmarks that better reflect the strengths of SNNs, such as temporal and event-driven tasks. 

Key issues such as identifying optimal SNN designs, understanding the impact of spike encoding and decoding, and scaling SNNs to handle larger and more complex datasets remain open challenges \cite{bouvier2019spiking_survey}. Addressing these challenges requires systematic exploration of the design space to develop efficient and scalable architectures. Given the computational constraints and temporal dynamics of SNNs, hardware-software co-design plays a critical role in ensuring that these architectures perform effectively on neuromorphic hardware \cite{aliyev2024_sparsityOverview}. This necessitates studies on NAS, which has emerged as a promising approach for automating the discovery of optimal SNN designs. By integrating NAS with hardware-aware techniques, researchers can identify architectures that balance accuracy, efficiency, and resource constraints, unlocking the full potential of SNNs for real-world applications.

\begin{table*}[!t]
\centering
\begin{threeparttable}
\setlength{\tabcolsep}{6pt}
\renewcommand{\arraystretch}{1.15}
\caption{Key Advantages and Challenges of SNNs}
\begin{tabular}{p{3.5cm} p{5.0cm} p{5.0cm}}
\toprule
\textbf{Feature} 
& \textbf{Advantages} 
& \textbf{Challenges} \\
\midrule

\textbf{Energy Efficiency} 
& Highly energy-efficient due to sparse and event-driven computation. 
& Simulation of multiple timesteps increases computational complexity. \\
\midrule

\textbf{Temporal Dynamics} 
& Excellent for time-dependent tasks (e.g., audio or sensory input). 
& Lack of standardized tools and methods for representing temporal data. \\
\midrule

\textbf{Biological Plausibility} 
& Mimics biological neurons, offering insights into neuroscience and fault-tolerant systems. 
& Training non-differentiable spike-based mechanisms remains challenging for gradient-based methods \cite{schuman2020evolutionary}. \\
\midrule

\textbf{Real-Time Processing} 
& Ideal for edge computing and IoT use cases requiring low-latency response. 
& Event-driven hardware platforms are still emerging and lack widespread adoption. \\
\midrule

\textbf{Scalability} 
& Potential to scale with neuromorphic hardware for efficient parallel processing. 
& Large datasets and deep architectures drive up computational costs, limiting scalability. \\
\midrule

\textbf{Training} 
& Surrogate gradient, conversion, and EO techniques enable learning despite spiking non-differentiability.
& Deep multi-layer SNNs remain difficult to train effectively, needing further research. \\
\midrule

\textbf{Hardware Integration} 
&Multi-objective optimization produces smaller, power-efficient networks tailored to specific hardware constraints for better integration with neuromorphic chips (e.g., Loihi~\cite{liu2024unleashing}) and improved hardware utilization.
& Neuromorphic hardware often supports only limited neuron and synapse model diversity. High communication overhead at the "hardware/software boundary" can bottleneck training if a simulator is unavailable \cite{schuman2020evolutionary}.\\
\midrule

\textbf{Accuracy} 
& Competitive results on event-driven or sensor-based benchmarks. 
& Underperforms ANNs on traditional tasks (e.g., MNIST, CIFAR-10) due to sensor mismatches. \\
\bottomrule

\end{tabular}
\label{table:SNNs-Advantages-Challenges}
\end{threeparttable}
\end{table*}

\section{Neural Architecture Search} \label{sec:nas}
NAS automates the tedious manual process of neural network design by employing optimization techniques that identify and rank optimal architectures for a given task based on a fitness metric (e.g., accuracy or computational efficiency). This search for the optimal configuration can be accelerated by approximating the ranking without fully training every candidate network. A misranking can occur when approximate evaluation produces a different ordering than full training with temporal execution.

An early NAS approach by Stanley et al. \cite{stanley2002evolving} used evolutionary methods, inspiring numerous refinements. Although these advances have benefited ANNs, SNNs pose distinct challenges due to their event-driven processing and temporal dynamics, necessitating specialized architectural considerations. To fully exploit SNNs' potential, NAS methods must be adapted accordingly. This section explores why traditional NAS techniques designed for ANNs may be ineffective for SNNs and examines existing NAS approaches for SNNs, focusing on the three core components: search space, search strategy, and evaluation \cite{elsken2019neural} (Figure \ref{fig:roadmap}).

\subsection{Unique Challenges of Applying NAS to SNNs} \label{sec:nas_challenges}
SNNs' temporal dynamics and event-driven processing demand specialized NAS approaches distinct from conventional ANN methods. Several key challenges complicate this adaptation:

First, while standard NAS relies on gradient-based optimization, SNNs' discrete spike functions are inherently non-differentiable. Though surrogate gradient methods exist, they introduce approximation errors and computational overhead that impact large-scale architecture searches \cite{deng2023surrogate}. The prevalence of spike-based learning rules like STDP, distinct from gradient-based approaches, further complicates traditional NAS applications.

Second, SNNs require specialized performance metrics beyond standard ANN measures. These include spike timing precision, firing rates, and temporal dynamics \cite{deng2020rethinking}, necessitating evaluation functions that specifically address spiking behavior.

Third, neuromorphic hardware considerations introduce unique constraints such as limited precision and asynchronous computation \cite{javanshir2022advancements}. Traditional NAS methods that do not account for these constraints may yield architectures incompatible with neuromorphic platforms.

Addressing these challenges requires specialized techniques that incorporate spiking neuron models and suitable network topologies \cite{masters, yan2024efficient}. Solutions must balance temporal coding performance, energy efficiency, and hardware compatibility \cite{putra2025spikenas, yan2024efficient}, while employing training mechanisms adapted for spike-based computation \cite{kim2022neural, masters}.

\subsection{Search Space Design} \label{sec:search_space}
The search space for a NAS method is the set of feasible network configurations, typically specified by parameters such as the number and type of layers and the layer hyperparameters (e.g., kernel size, filter sizes, or the number of neurons). Because these dimensions yield billions of potential architectures, the search space can be prohibitively large and often becomes the primary bottleneck for architecture search performance \cite{pham2018efficient}. Moreover, different architectural choices can significantly impact hardware accelerators, affecting power efficiency, memory usage, and throughput \cite{aliyev2024fine,aliyev2024_sparsityOverview}. Consequently, it is important to design the search space to balance performance and practical hardware constraints.

The search can be simplified and the search space reduced by incorporating prior knowledge of well-performing architectural patterns. However, doing so may introduce human bias, undermining the discovery of novel solutions. On the other hand, overly constraining the search space might exclude high-performing architectures, resulting in suboptimal performance. Hence, selecting an appropriate search space remains a critical challenge in SNNaS, requiring consideration of both algorithmic and hardware factors.

Current NAS methods use a variety of search spaces.

\paragraph{Global (Unstructured) Search Spaces} Global search spaces consider the entire network configuration at once, allowing architectural decisions to be made across all layers simultaneously without enforced repetition or modular reuse. They offer tremendous diversity but can be extremely large, which makes efficient exploration challenging. Several SNNaS methods employ global search spaces to maximize architectural flexibility. For instance, MixedSNN~\cite{xie2023efficient} searches among different spiking neuron models and thresholds, and Auto-Spikformer~\cite{che2023auto} combines transformer and SNN parameters. Methods such as Liu and Yi~\cite{liu2024unleashing}, efficient SNN (ESNN)~\cite{yan2024efficient}, Automatic Spiking Transformer (AutoST)~\cite{wang2024autost}, Genetic Algorithm-based Quantization framework for Spiking Neural Networks (GAQ-SNN)~\cite{gaq}, and asynchronous neuromorphic hardware architecture search (ANAS)~\cite{zhang2023anas} illustrate the complexity of global search—spanning from trillions of spiking configurations to hardware-related parameters like processing elements and buffer sizes.

\paragraph{Sequential Search Spaces} Inspired by early NAS works~\cite{zoph2017neural,baker2016designing}, sequential search spaces encode neural architectures as ordered, layer-by-layer constructions, where each layer is specified as an individual architectural choice (e.g., operation type, neuron model, or kernel size), rather than through repeated modules or hierarchical architectural motifs. Because these spaces do not enforce architectural regularity or parameter sharing, their combinatorial size grows rapidly, substantially increasing search complexity. In spiking neural networks, this challenge is further amplified by temporal dynamics, neuron state variables, and spike-driven computation, rendering purely sequential SNNaS prohibitively expensive in practice. Consequently, sequential search spaces remain largely unexplored in practical SNNaS settings, with most modern methods favoring more structured alternatives that better control computational cost.

\begin{figure}[t!]
		\centering
		\includegraphics[width=\linewidth]{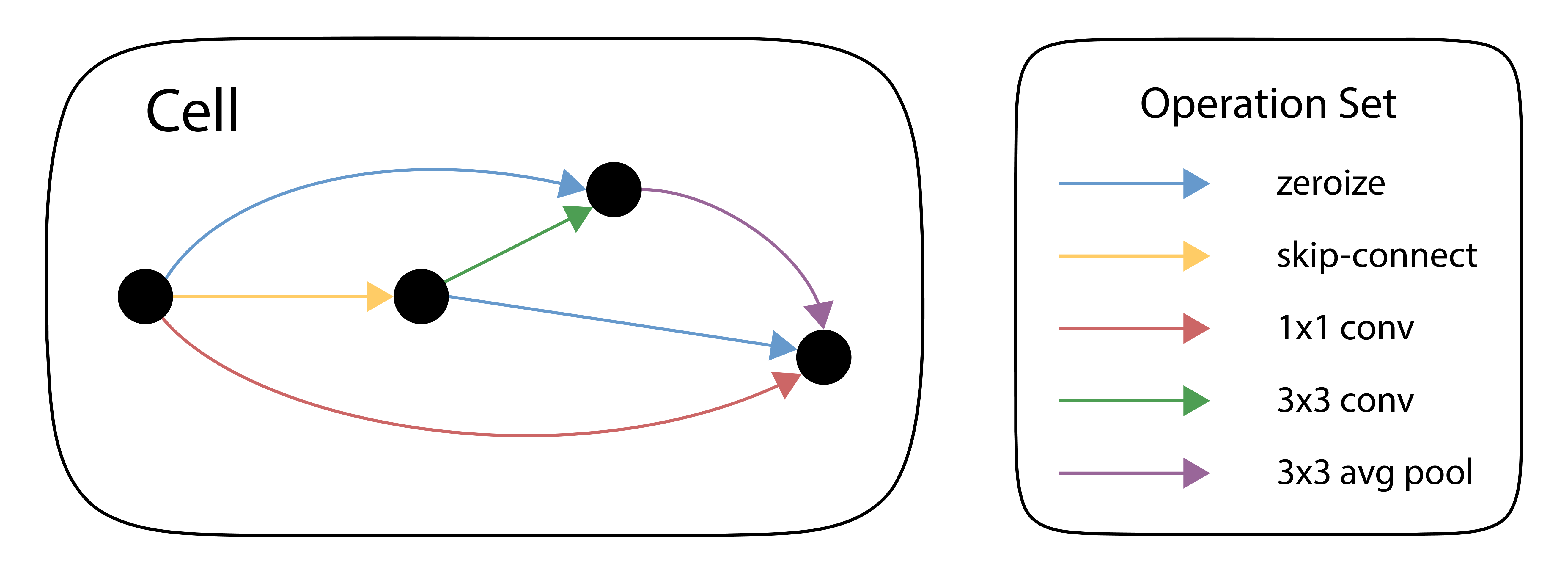}
		\caption{Example of cell with four nodes and five possible operations. The cell is a directed acyclic graph, with each edge representing an operation \cite{dong2020bench}.}	\label{fig:cell-based}
\end{figure}

\paragraph{Cell-Based Search Spaces} Cell-based approaches define networks at macro and micro levels. Macro-architectures arrange repeated cells, while micro-architectures define operations within each cell. SNASNet (NAS approach for finding better SNN architectures)~\cite{kim2022neural}, based on NAS-Bench-201~\cite{dong2020bench}, pioneered cell-based SNN search using directed acyclic graphs (DAGs) of operations (zeroize, skip connection, $1\times1$ convolution, $3\times3$ convolution, and $3\times3$ average pooling). Extensions include spiking spatiotemporal NAS (SSTNAS)~\cite{li2024spiking}, SpikeNAS~\cite{putra2025spikenas}, Spikernel~\cite{putra2024methodology}, NeuroNAS~\cite{putra2024neuronas}, and variations in cell design \cite{masters}. SpikeDHS (spike-based differentiable hierarchical search)~\cite{che2022differentiable, che2024spatial} also employs DAG-based cells. Multi-Attention Differentiable Architecture Search (MA-DARTS)~\cite{man2024differentiable} extends the cell-based framework to incorporate multi-dimensional attention blocks that distinguish information across channel, spatial, and temporal dimensions and prioritize spatiotemporal spike patterns. MA-DARTS has minimal parameter overhead and improved accuracy. An example cell-based diagram is shown in Figure~\ref{fig:cell-based}.

\begin{figure*}[t!]
		\centering
		\includegraphics[width=0.75\linewidth]{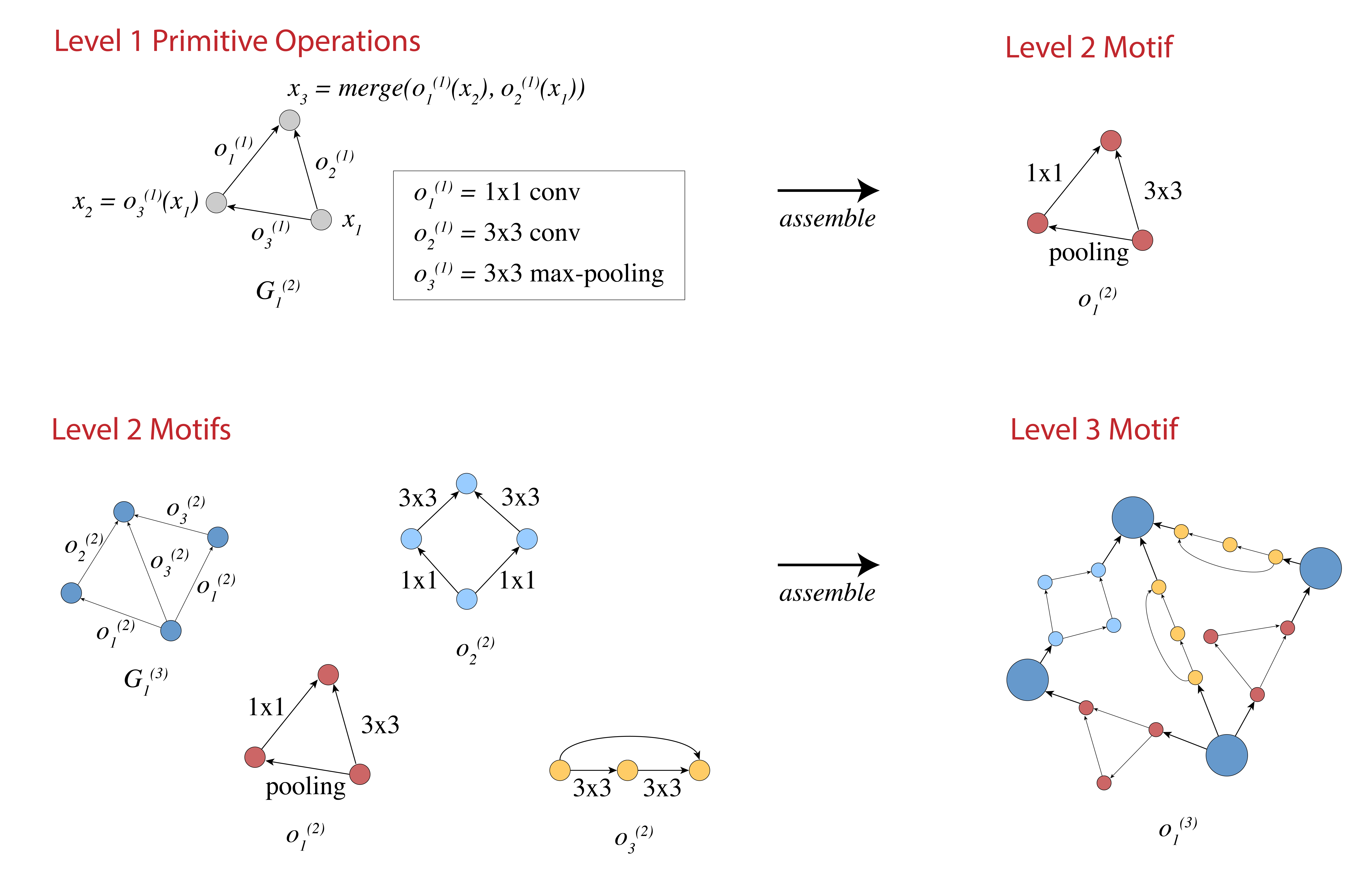}
		\caption{Example of a three-level hierarchy taken from Liu et al.~\cite{liu2018hierarchical}. Level-1 primitive operations $o_1^{(1)}$, $o_2^{(1)}$, and $o_3^{(1)}$, representing $1 \times 1$ convolution, $3 \times 3$ convolution, and $3 \times 3$ max-pooling respectively, are assembled into a level-2 motif, $o_1^{(2)}$ (TOP). Level-2 motifs $o_1^{(2)}$, $o_2^{(2)}$, and $o_3^{(2)}$ are then assembled into a level-3 motif $o_1^{(3)}$ (BOTTOM).}	\label{fig:hierarchy}
\end{figure*}

\paragraph{Hierarchical Structure Search Spaces} Hierarchical NAS generalizes cell-based approaches and arranges architectures in multi-level motifs, mirroring modular designs like VGGNet~\cite{simonyan2015deep} or ResNet~\cite{he2016deep}. Multiscale evolutionary NAS (MSE-NAS)~\cite{pan2023multi} organizes SNNs into microscopic, mesoscopic, and macroscopic levels; AutoSNN~\cite{na2022autosnn} and Asynchronous Neuromorphic algorithm/hardware Co-Exploration Framework (ANCoEF)~\cite{zhang2024ancoef} combine macro-level backbones with micro-level spiking blocks. SpikeDHS~\cite{che2022differentiable, che2024spatial} extends from cell-level to layer-level search, and LitE-SNN~\cite{liu2024lite} adds pruning and mixed-precision quantization. SpikeExplorer~\cite{padovano2024spikeexplorer} employs Bayesian optimization (BO) for multi-objective search. Deep Evolutionary SPIking NEtwork (DESPINE)~\cite{ajay2024despine} and Neural circuit Evolution strategy (NeuEvo)~\cite{shen2023brain} further emphasize biologically inspired motifs, excitatory/inhibitory neurons, and feedback connections. Figure~\ref{fig:hierarchy} illustrates a typical three-level hierarchy.

\paragraph{Hardware Design Search Spaces} As network architectures directly affect implementation efficiency (e.g., power, memory), hardware aspects must be considered early in SNNaS. Parameter-based frameworks~\cite{abdelfattah2020best,chen2020you,zhang2024ancoef} adjust processing elements (PEs), buffer sizes, and interfaces, whereas template-based methods~\cite{yang2020co} reconfigure known hardware design templates. ANCoEF~\cite{zhang2024ancoef} highlights that incorrect neuron-to-PE ratios can waste bits in lookup tables (LUTs), weight SRAM, and network-on-chip flits. Benmeziane et al.~\cite{benmeziane2021comprehensive} classify hardware into server-grade CPUs/GPUs/FPGAs/ASICs, mobile devices, and tiny devices—each requiring distinct design optimizations. When jointly explored, neural and hardware search spaces enable end-to-end optimization of PPA metrics (power, performance, and area) and accuracy at the cost of increased search complexity.

\begin{table*}[ht!]
\centering
\begin{threeparttable}
\setlength{\tabcolsep}{8pt}
\renewcommand{\arraystretch}{1.15}
\caption{Hardware Constraints and Optimization Techniques}
\label{table:hw-optimizations}
\begin{tabular}{p{3.0cm} p{4.5cm} p{6.5cm}}
\toprule
\textbf{Hardware Constraint} &\textbf{Description} &\textbf{Examples of Optimization Techniques}\\
\midrule
\textbf{Memory Usage}
&The amount of memory required to store model parameters and operations.
& Mixed-precision quantization (e.g., LitE-SNN~\cite{liu2024lite}, ABC-Arc~\cite{zhang2025spike}, NeuroNAS~\cite{putra2024neuronas}, ASNPC~\cite{wang2025asnpc}); 
Memory-aware kernel size exploration (e.g., SpikeNAS~\cite{putra2025spikenas}, NeuroNAS~\cite{putra2024neuronas}); 
Weight pruning and compression (e.g., GAQ-SNN~\cite{gaq}, ABC-Arc~\cite{zhang2025spike}, SPEAR~\cite{xie2025spear}, ESL-SNNs~\cite{shen2023esl}, ANCoEF~\cite{zhang2024ancoef}); Sparse training from scratch (e.g., ESL-SNNs~\cite{shen2023esl}); Linear regression for SynOps estimation (e.g., SPEAR~\cite{xie2025spear}); DRL target-aware reward (e.g., SPEAR~\cite{xie2025spear}); Neuron hardware unit reuse (e.g., ASNPC~\cite{wang2025asnpc}).\\

\midrule
\textbf{Latency}&Time taken to complete inference or training, crucial for real-time applications.& Hardware-specific optimization (e.g., NeuroNAS~\cite{putra2024neuronas}, ANCoEF~\cite{zhang2024ancoef}, ASNPC~\cite{wang2025asnpc}, SpikeExplorer~\cite{padovano2024spikeexplorer}, TNAS-ER~\cite{liu2025efficient}); Timestep compression in SNNs (e.g., LitE-SNN~\cite{liu2024lite}, DESPINE~\cite{ajay2024despine}, TNAS-ER~\cite{liu2025efficient}); Sparsity-aware event skipping (e.g., TNAS-ER~\cite{liu2025efficient}); ANN-assisted time-to-first-spike (TTFS) search (e.g., SpikeExplorer~\cite{padovano2024spikeexplorer}).
\\

\midrule
\textbf{Energy Consumption}&The power required to perform computations, important for low-power devices.& Minimizing spike counts (e.g., NeuEvo~\cite{shen2023brain}, AutoSNN~\cite{na2022autosnn}, DESPINE~\cite{ajay2024despine}, EB-NAS~\cite{pan2024brain}, ANCoEF~\cite{zhang2024ancoef}, MONAS-ESNN~\cite{saghand2025monas}); Synaptic operation (SynOps)-based evaluation (e.g., Auto-Spikformer~\cite{che2023auto}, ESNN~\cite{yan2024efficient}, SPEAR~\cite{xie2025spear}, LitE-SNN~\cite{liu2024lite}); Sparse computation using event-driven architectures (e.g., SpikeX~\cite{xu2025spikex}); Linear regression for SynOps estimation (e.g., SPEAR~\cite{xie2025spear}); DRL target-aware reward (e.g., SPEAR~\cite{xie2025spear}); Sparsity-aware weight loading (e.g., SpikeX~\cite{xu2025spikex}); ASAHD (e.g., MONAS-ESNN~\cite{saghand2025monas}).\\
\midrule

\textbf{Hardware Area}&The physical size of the hardware components required for implementation.& Compute-in-memory architectures (e.g., NeuroNAS~\cite{putra2024neuronas}); Optimization of neuron and synapse allocations (e.g., ANCoEF~\cite{zhang2024ancoef}, ANAS~\cite{zhang2023anas}, MOO-SNNs~\cite{dimovska2019multi}); Neuron hardware unit reuse (e.g., ASNPC~\cite{wang2025asnpc}); Mixed-precision quantization (e.g., GAQ~\cite{gaq}).\\
\midrule

\textbf{Hardware Compatibility}&Ensures the designed SNN can efficiently operate on specific hardware.& Co-exploration of architecture and hardware (e.g., ANAS~\cite{zhang2023anas}, ANCoEF~\cite{zhang2024ancoef}, ASNPC~\cite{wang2025asnpc}, NeuroNAS~\cite{putra2024neuronas}, SpikeExplorer~\cite{padovano2024spikeexplorer}); Resource-balancing algorithms for neuromorphic platforms (e.g., ANAS~\cite{zhang2023anas}, ANCoEF~\cite{zhang2024ancoef}, TNAS-ER~\cite{liu2025efficient}); Chip-specific search space (e.g., Liu \& Yi~\cite{liu2024unleashing}).\\
\midrule

\textbf{Bandwidth}&Data transfer capacity between components, critical in parallel architectures.& Data partitioning and mapping (e.g., ANCoEF~\cite{zhang2024ancoef}, ANAS~\cite{zhang2023anas}, SpikeX~\cite{xu2025spikex}, TNAS-ER~\cite{liu2025efficient}); Efficient routing and scheduling strategies (e.g., ANAS~\cite{zhang2023anas}, SpikeX~\cite{xu2025spikex}); Compute-in-memory architectures (e.g., NeuroNAS~\cite{putra2024neuronas}); Sparsity-aware weight loading (e.g., SpikeX~\cite{xu2025spikex}); Activity-based data fetching prioritization (e.g., SpikeX~\cite{xu2025spikex}).\\
\bottomrule

\end{tabular}
\end{threeparttable}
\end{table*}

\paragraph{Neuron-Level Search Spaces} 
In SNNaS, neuron-level search spaces optimize spiking neurons to suppress redundant spiking activity, ensuring evolved architectures are more accurate and energy-efficient. While some SNNaS works select from different neuron model types tailored to their task~\cite{xie2023efficient, zhangspatio}, others optimize neuron parameters that might include leakage, decay, spiking frequencies, adaptive thresholds, time steps, reset mechanisms, and even surrogate gradient shapes to better capture temporal dynamics \cite{ajay2024despine, yan2024efficient, che2022differentiable, che2024spatial, che2023auto, schuman2016evolutionary, branquinho2023spenser, zhou2020surrogate, pan2023multi, wang2025asnpc}.

\subsection{Hardware-Aware NAS} \label{sec:hw_nas}
Hardware-aware NAS (HW-NAS) seeks to optimize neural network architectures for specific fixed target hardware instantiations (e.g., GPUs, FPGAs, mixed-signal neuromorphic hardware, asynchronous platforms, and emerging hybrid architectures), balancing traditional performance metrics (e.g., accuracy) with hardware-related constraints like energy, latency, memory usage, and computational requirements, without altering the underlying neural architecture search space \cite{zhang2019neural,benmeziane2021comprehensive}. Importantly, temporal and event-driven SNN computations behave differently across hardware implementations~\cite{pfeiffer2018deep}, so even theoretically optimal SNNs may be impractical if they fail to align with platform-level design choices. As summarized in Table~\ref{table:hw-optimizations}, HW-NAS ensures that the final architectures exploit hardware features---like event-driven processing or sparse computation~\cite{rathi2023exploring}---to achieve substantial efficiency gains. 

A variety of accelerators support SNNs, ranging from GPGPU systems \cite{qu2020high, yavuz2016genn, beyeler2015carlsim} and FPGA-based designs \cite{han2020hardware, pani2017fpga, liu2022fpga, ju2020fpga, ma2017darwin, li2021fast} to neuromorphic platforms like Loihi~\cite{davies2018loihi} and TrueNorth~\cite{truenorth}, mixed-signal VLSI \cite{hsieh2012vlsi, schemmel2010wafer}, and nanotechnology \cite{gao2007cortical}. Although neuromorphic hardware can be highly energy-efficient, it often supports only a subset of neuron and synaptic models; more flexible general-purpose computing systems can simulate diverse designs but may consume more resources \cite{painkras2013spinnaker}. Thus, HW-NAS must consider whether the targeted hardware is specialized (e.g., tailored to a certain neuron model) or general-purpose (e.g., a GPU).

In general, there are two main strategies for HW-NAS: single-objective and multi-objective optimization \cite{benmeziane2021comprehensive}. Single-objective methods either use a two-stage process---first searching for accuracy, then optimizing for hardware---or incorporate hardware limits as constraints from the onset. Multi-objective frameworks, such as SpikeExplorer~\cite{padovano2024spikeexplorer} and LitE-SNN~\cite{liu2024lite}, balance accuracy with hardware metrics (e.g., power, area, and memory). Many SNNaS methods rely on Synaptic Operations (SynOps) to model computational and energy costs~\cite{che2023auto,liu2024lite,gaq,yan2024efficient}, while spike-count minimization (e.g.,\cite{na2022autosnn,pan2024brain,shen2023brain}) reduces power. Quantization~\cite{gaq} and kernel-size optimization~\cite{putra2024methodology} further shrink memory footprints. HW-NAS methods like NeuroNAS~\cite{putra2024neuronas}, NeuEvo~\cite{shen2023brain}, GAQ-SNN~\cite{gaq}, and Auto-Spikformer~\cite{che2023auto} show that carefully coupling architecture search with hardware-specific constraints yields high accuracy alongside notable improvements in energy and resource efficiency.

\subsection{Bridging the Hardware-Software Co-Design Gap in SNNaS}

\subsubsection{Co-Exploring the Neural Architecture Search Space and Hardware Design Space} \label{sec:coexploration}

Co-exploration simultaneously searches neural architectures and hardware design parameters, going beyond standard HW-NAS frameworks by optimizing every dimension of the design jointly~\cite{jiang2020standing}. Unlike HW-NAS, which assumes a fixed hardware platform, co-exploration may consider different or new hardware architectures to ensure designs are feasible, efficient, and well-aligned with real-world deployment constraints~\cite{zhang2019neural}. This approach typically delivers better performance and lower costs than separately optimizing hardware or SNNs, but at the expense of greater search complexity, substantial GPU hours, and a large carbon footprint~\cite{strubell2019carbon}.

ANAS~\cite{zhang2023anas} exemplifies co-exploration by using an evolutionary algorithm to search a comprehensive design space, evaluating hardware performance with the Configurable Asynchronous Neuromorphic hardware simulator (CanMore). Compared to random or grid search, ANAS achieves up to an order-of-magnitude improvement in energy-delay and is four orders of magnitude faster to search. Similarly, ANCoEF~\cite{zhang2024ancoef} combines a deep reinforcement learning (DRL)- based multi-objective hardware search with a supernet-based SNN search~\cite{na2022autosnn}, partially training candidate architectures and discarding those that fail hardware constraints. This iterative process yields a 1.81$\times$ reduction in energy-delay product (EDP) and a 2.73$\times$ speedup over ANAS~\cite{zhang2023anas} on the N-MNIST dataset, highlighting the promise of fully joint NAS and hardware optimization.

Studies show neuromorphic hardware-software co-design achieves breakthrough performance in narrow, well-defined domains where end-to-end optimization is possible \cite{zhang2024ancoef, xu2025spikex, wang2025asnpc, ajay2024despine}, but fails systematically when architectural misalignment occurs or communication bottlenecks are present \cite{buettner2021case, balaji2020compiling, pfeiffer2018deep, seekings2024towards}.

\subsubsection{The Hardware/Software Boundary} \label{sec:hw_sw_boundary}

Perhaps the most critical frontier in SNNaS is the "hardware/software boundary," the interface where mathematical abstractions must reconcile with the physical realities of neuromorphic substrates \cite{deng2024spiking, schuman2020evolutionary, putra2024neuronas}. Standard NAS approaches typically treat hardware as a fixed penalty term in a fitness function, but this misses a crucial insight: synchronous (clock-driven) and asynchronous (event-driven) hardware require fundamentally different architectural paradigms \cite{deng2024spiking}. Performance is often further constrained by communication overhead between software simulators and physical chips \cite{schuman2020evolutionary}.

\subsubsection{Asynchronous and Event-Driven Hardware Constraints}

Current asynchronous, event-driven hardware platforms like Intel's Loihi and Synsense's Speck achieve their power efficiency by eliminating the global clock, but this design choice imposes stringent architectural constraints \cite{deng2024spiking, liu2024unleashing}. Without clock synchronization, spike arrival times become imprecise, making certain operations problematic for existing asynchronous neuromorphic chips. Spiking matrix multiplication (as used in Spiking Transformers) and max-pooling layers are particularly vulnerable—even small timing delays on contemporary asynchronous neuromorphic platforms can produce significant errors in attention maps \cite{deng2024spiking}. Hardware-specific limitations compound these challenges; Loihi, for instance, lacks support for backward and shortcut connections \cite{liu2024unleashing}.

The Spiking Token Mixer (STMixer)~\cite{deng2024spiking} navigates these constraints by restricting its search to asynchronous-compatible operations—convolutions, fully connected layers, and residual paths—while deliberately avoiding operations that demand precise spike synchronization. ANAS~\cite{zhang2023anas} tackles the problem from the hardware side, using CanMore to search across both numerical parameters (PE count, buffer size) and structural choices (NoC topologies, routing algorithms), optimizing how Address-Event Representation protocols and axon-sharing rules are implemented to match SNN sparsity patterns.

\begin{table*}[ht!]
\centering
\caption{Spiking Neural Network Architecture Search Strategies}
\label{nas-methods}
\begin{threeparttable}
\setlength{\tabcolsep}{6pt}
\renewcommand{\arraystretch}{1.15}

\setlength{\tabcolsep}{5pt}
\renewcommand{\arraystretch}{1.15}

\begin{tabularx}{\textwidth}{
>{\raggedright\arraybackslash}p{2.2cm} 
>{\raggedright\arraybackslash}p{3.2cm}  
Y 
Y 
>{\raggedright\arraybackslash}p{2.8cm}   
}

\toprule
\textbf{Search Method} 
& \textbf{Works} 
& \textbf{Key Characteristics} 
& \textbf{Strengths} 
& \textbf{Limitations} \\
\midrule

\textbf{Grid Search} 
& NeuroNAS \cite{putra2024neuronas}, ANAS~\cite{zhang2023anas}, \cite{abderrahmane2020design}
& Systematically evaluates all hyperparameter combinations.
& Parallelizable and simple to implement. Ensures the optimum architecture is found.
& Computationally intensive. Impractical for large search spaces. \\
\midrule

\textbf{Random Search} 
& SNASNet \cite{kim2022neural}, Liu \& Yi \cite{liu2024unleashing}, Spikernel~\cite{putra2024methodology}, STMixer~\cite{deng2024spiking}, ESL-SNNs~\cite{shen2023esl}, ANAS~\cite{zhang2023anas}\tnote{1}, ESNN~\cite{yan2024efficient}\tnote{1}, TNAS-ER~\cite{liu2025efficient}\tnote{1}, Auto-Spikformer~\cite{che2023auto}\tnote{2}, AutoSNN~\cite{na2022autosnn}\tnote{2}
& Selects random combinations of hyperparameters from the search space and evaluates the resulting architecture performances.
& Explores the search space broadly, finds good configurations faster, and often achieves near-optimal solutions with fewer evaluations. Simple to implement and parallelizable.
& Lacks guided exploration. Does not adaptively focus on promising regions of the search space. \\
\midrule

\textbf{Evolutionary}
&AutoSNN~\cite{na2022autosnn}, GAQ~\cite{gaq}, Auto-Spikformer~\cite{che2023auto}, NeuEvo~\cite{shen2024evolutionary}, MixedSNN~\cite{xie2023efficient}, ANAS~\cite{zhang2023anas}, MSE-NAS~\cite{pan2023multi}, DESPINE~\cite{ajay2024despine}, AutoST~\cite{wang2024autost}, SSTNAS~\cite{li2024spiking}, EONS~\cite{schuman2016evolutionary}, ABC-Arc~\cite{zhang2025spike}, EMO-SNAS~\cite{song2025evolutionary}, EB-NAS~\cite{pan2024brain}, PESNN~\cite{wang2025predictor}, SPENSER~\cite{branquinho2023spenser}, BCNAS-SNN~\cite{dengproper}, MONAS-ESNN~\cite{saghand2025monas}, GP-Assisted CMA-ES~\cite{zhou2020surrogate}, Membrane-NAS~\cite{liu2025auto}, Spiking WANN~\cite{anwar2021evolving}
& Iteratively selects, changes, and reproduces neural networks to optimize a fitness value related to an objective like classification accuracy.
& Does not require differentiable fitness functions. Suitable for multi-objective optimization.
& Computationally expensive with more parameters. \\
\midrule

\textbf{Deep Reinforcement Learning (DRL)} 
& ANCoEF \cite{zhang2024ancoef}, SPEAR~\cite{xie2025spear}, GAQ~\cite{gaq}
& Rewards positive behaviors and penalizes negative ones to derive efficient strategies toward a given goal. Balances exploration and exploitation.
& Scales to complex and high-dimensional search spaces. Flexible reward design enables custom multi-objective optimization. 
& Computationally expensive and sample-inefficient. Compounded latency with SNNs. Training instability and tuning sensitivity. Additional controller infrastructure required. Increased latency in SNN settings. \\
\midrule

\textbf{Bayesian Optimization (BO)} 
& SpikeExplorer \cite{padovano2024spikeexplorer}, ASNPC~\cite{wang2025asnpc}, GP-Assisted CMA-ES~\cite{zhou2020surrogate}, Benmeziane et al.~\cite{benmeziane2023skip}, deepHyper~\cite{yanguas2022automl}
& Finds the optimum of expensive and poorly understood functions using a probabilistic model-based approach.
& Requires fewer evaluations. Capable of finding global optima in complex, multi-modal search spaces. Learns from prior evaluations to focus on promising regions.
& Not well-suited for high-dimensional spaces. Computationally intensive and difficult to parallelize. Can converge to local optima in noisy settings. \\
\midrule

\textbf{Gradient-Based} 
& SpikeDHS~\cite{che2022differentiable, che2024spatial}, ESNN~\cite{yan2024efficient}, LitE-SNN~\cite{liu2024lite}, MA-DARTS~\cite{man2024differentiable}, ST-DANO~\cite{zhangspatio}, SpikeDHS~\cite{che2022differentiable, che2024spatial}
& Allows for continuous relaxation of the architecture space, enabling gradient-based optimization.
& Fast exploration, scalable to large spaces. Efficient for complex architecture searches.
& May under-explore the search space. Susceptible to local optima. \\
\bottomrule

\end{tabularx}
\begin{tablenotes}
\item[1] Implemented in comparative baseline.
\item[2] Implemented in ablation studies.
\end{tablenotes}
\end{threeparttable}
\end{table*}

\subsection{Neural Network Architecture Search Strategies} \label{sec:search_strategies}
NAS involves exploring a vast design space to identify high-performing configurations for a specific task while minimizing computational resources and time. A successful strategy must evaluate performance (often by training on a reduced dataset subset), estimate hardware cost (via real-world measurement or modeling), and select an appropriate search algorithm (e.g., grid search, random search, evolutionary algorithms, deep reinforcement learning, Bayesian optimization, or gradient-based methods) \cite{benmeziane2021comprehensive}. Here we briefly describe different popular NAS strategies, with an emphasis on their applicability to SNNaS, as summarized in Table \ref{nas-methods}.

\subsubsection{Grid Search}
Grid search systematically explores all hyperparameter combinations but becomes impractical as search spaces grow exponentially~\cite{liashchynskyi2019grid}. NeuroNAS~\cite{putra2024neuronas}, an SNNaS framework, adapts the grid search by pruning low-impact SNN operations and incorporating hardware constraints directly, achieving \(8.9\times\) speedup over SNASNet~\cite{kim2022neural} while maintaining accuracy on CIFAR and TinyImageNet.

\subsubsection{Random Search}
Random search samples configurations at random, often discovering novel designs missed by structured methods~\cite{bergstra2012random,yu2019evaluating,li2020random,radosavovic2020designing}. SNASNet~\cite{kim2022neural} uses initialization-time, training-free (no weight updates) spike activation diversity metrics to identify promising random SNN topologies. Liu et al.~\cite{liu2024unleashing} randomly sampled 1,000 architectures for Loihi, and trained the top 20 according to gradient metrics to achieve competitive accuracy with lower energy than state-of-the-art SNNs.

\begin{figure}[!t]
		\centering
		\includegraphics[width=0.5\textwidth]{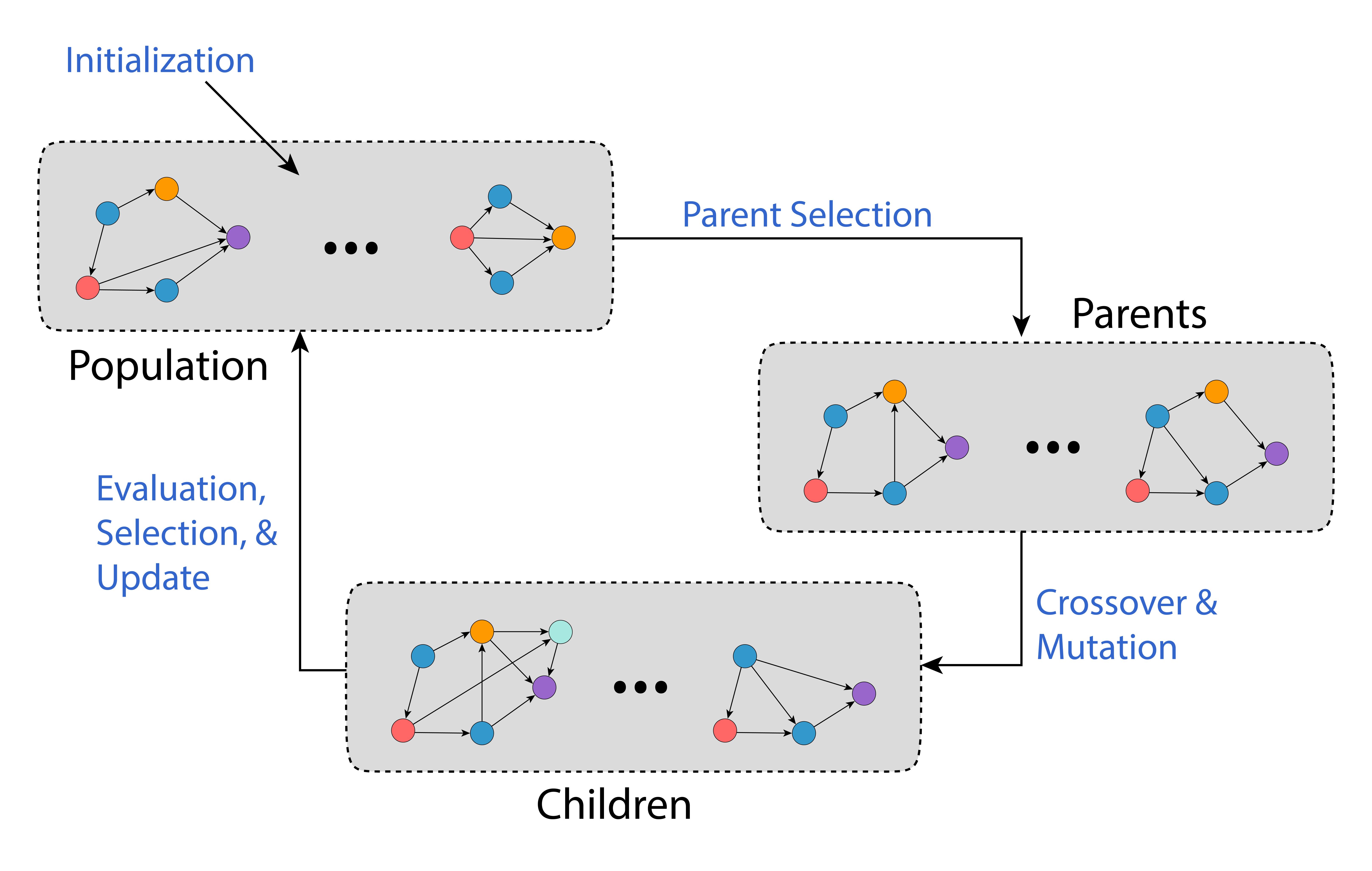}
		\caption{Evolutionary Algorithm for NAS based on \cite{ren2021comprehensive}. After initializing the population, the algorithm selects parents and then crosses or mutates them to generate children. It evaluates the adaptability, or the neural architecture performance, of the children and selects a group of individual neural architectures with the best performance from the children generated.}	\label{fig:evolution}
\end{figure}

\subsubsection{Evolutionary Methods} Evolutionary NAS methods (Figure~\ref{fig:evolution}) are rooted in early works \cite{miller1989designing,guha1988genetic,todd1988evolutionary} that mimic biological evolution by iteratively selecting, mutating, and recombining neural network configurations to optimize a fitness function---often accuracy, energy usage, or other objectives \cite{liu2021survey, schuman2016evolutionary}. Unlike gradient-driven approaches, they do not require a differentiable objective, making them suitable for multi-objective and discrete design challenges~\cite{schuman2016evolutionary}. Genetic Algorithms (GAs) are a common technique that encode networks as strings or graphs, then repeatedly apply crossover and mutation to discover high-performing architectures~\cite{schuman2016evolutionary, ajay2024despine, branquinho2023spenser}. Historically, these methods have optimized weights (e.g., NeuroEvolution of Augmenting Topologies (NEAT)~\cite{stanley2002evolving}), and more recent works extend them to structural and hyperparameter searches \cite{shrestha2019optimizing, young2015optimizing, real2017large}, sometimes using illumination algorithms like Multi-dimensional Archive of Phenotypic Elites (MAP-Elites)~\cite{mouret2015illuminating} for diverse solutions.

Despite high computational demands (e.g., AmoebaNet-A~\cite{real2019regularized} in ANN NAS required 3,150 GPU days), evolutionary algorithms can be accelerated through smaller training sets, prediction models, Lamarckian strategies \cite{prellberg2018lamarckian, elsken2018efficient}, or partial weight inheritance. Notable ANN HW-NAS efforts include TinyNAS~\cite{lin2020mcunet}, Once-for-all~\cite{cai2020once}, Hardware-Aware Transformers (HAT)~\cite{wang2020hat}, NAS for convolutional capsule networks (NASCaps)~\cite{marchisio2020nascaps}, and HURRICANE~\cite{zhang2020fast}. For SNNs, evolutionary approaches are similarly powerful for tasks requiring hardware, memory, and energy optimizations, as demonstrated by EONS~\cite{schuman2016evolutionary}, GAQ-SNN~\cite{gaq}, DESPINE~\cite{ajay2024despine}, MixedSNN~\cite{xie2023efficient}, MSE-NAS~\cite{pan2023multi}, Auto-Spikformer~\cite{che2023auto}, NeuEvo~\cite{shen2023brain}, and ANAS~\cite{zhang2023anas}, among others. Their parallelizable, global search can reveal novel SNN designs in scenarios where gradient signals are limited, although balancing computation, efficient encoding, and hyperparameter tuning remains a key challenge.

\subsubsection{Reinforcement Learning Methods}
Reinforcement learning (RL) methods, illustrated in Figure \ref{fig:reinforcement}, frame NAS as a Markov Decision Process \cite{sutton2018reinforcement}, where an agent explores network architectures (or hardware designs) by taking actions that yield rewards based on accuracy or efficiency. In practice, most RL-based NAS methods effectively employ deep reinforcement learning (DRL), where policies are parameterized by neural networks to manage high-dimensional, discrete design spaces. Early work by Zoph and Le \cite{zoph2017neural} introduced DRL-based controllers for NAS but required extensive computational resources; subsequent techniques, such as weight sharing \cite{pham2018efficient} and learning curve prediction \cite{baker2017accelerating} reduced this overhead. DRL is appealing for hardware-oriented NAS because controllers can dynamically adjust to multi-objective goals (e.g., latency, energy)~\cite{zhang2019neural,tan2019mnasnet,lu2019neural,abdelfattah2020best,jiang2020standing}.

In the SNNaS context, ANCoEF~\cite{zhang2024ancoef} demonstrates a multi-objective DRL approach that simultaneously searches SNN architectures (via a supernet-based method \cite{na2022autosnn}) and asynchronous hardware parameters (e.g., neuron connectivity, mapping, partitioning, arbitration). Its DRL agent selects actions to maximize a reward reflecting accuracy, latency, energy consumption, and area utilization. By effectively modeling non-numerical design options (e.g., routing strategies) as discrete actions, ANCoEF outperforms the evolutionary algorithm-based ANAS~\cite{zhang2023anas} on N-MNIST with a 1.81$\times$ lower EDP and 2.73$\times$ less search time.

Despite these advances, DRL-based SNNaS methods are relatively unexplored. Challenges include high computational demands, large action spaces, potential local optima, and dynamic SynOps \cite{liu2021survey, ren2021comprehensive, song2025evolutionary, xie2025spear}. DRL controllers can be sample-inefficient and may require extensive tuning to generalize across tasks \cite{xie2025spear}. Future directions involve more robust designs, refined reward functions, and possible hybridization with evolutionary or BO methods to mitigate sample intensity and improve scalability.

\begin{figure}[t]
		\centering
		\includegraphics[width=0.5\textwidth]{figures/rl2.png}
		\vspace{-10pt}
		\caption{Illustration of RL for NAS based on \cite{zoph2017neural}. }	\label{fig:reinforcement}
\end{figure}

\subsubsection{Bayesian Optimization}
BO adaptively searches expensive, complex objective functions by building a probabilistic surrogate model (e.g., a Gaussian process) and balancing exploration and exploitation via an acquisition function~\cite{shahriari2015taking}. In SNNaS, BO typically encodes architectural elements (e.g., neuron types, connectivity) and performance metrics (e.g., accuracy, sparsity) into a surrogate model that guides the next candidate selection, thereby minimizing costly simulation and training runs. The approach iteratively updates the surrogate model with newly evaluated architectures, refining predictions and uncertainties; careful tuning of hyperparameters, surrogate design, and acquisition functions remains essential. Despite potential computational overhead and risks of local optima, BO’s data-efficient search can yield high-quality solutions, as shown by SpikeExplorer~\cite{padovano2024spikeexplorer}, which uses BO to co-optimize power, latency, area, and accuracy on FPGAs. This enables efficient exploration across multiple constraints, ultimately outperforming several accelerator baselines while preserving accuracy.

\subsubsection{Gradient-Based Methods}
Gradient-based NAS methods optimize architecture parameters via backpropagation by continuously relaxing discrete choices~\cite{liu2018darts}. Early examples include Differentiable Architecture Search (DARTS)~\cite{liu2018darts}, Stochastic NAS (SNAS)~\cite{xie2018snas}, ProxylessNAS~\cite{proxyless}, and FBNet (a hardware-aware efficient ConvNet design via differentiable NAS)~\cite{wu2019fbnet}, which reduced the memory overhead and search time by restricting continuous activation paths or using stochastic gradient updates. However, some studies, such as Bingham et al.~\cite{binghamineffectiveness}, report that gradient descent may stagnate on a single architecture over many iterations, failing to effectively exploit the broader search space as random or evolutionary methods do. Despite these challenges, gradient-based strategies remain popular due to their efficiency once continuous relaxation is established, and they are frequently combined with techniques such as inference latency constraints, as in HTAS~\cite{jiang2020hardware}.

Gradient-based methods are common in SNNaS. SpikeDHS~\cite{che2022differentiable} adapts the DARTS~\cite{liu2018darts} framework for SNNs by implementing spike-based computation at cell and layer levels, ensuring multiplication-free inference. It uses Differentiable Surrogate Gradient Search (DGS) to optimize local gradient approximations and prevent the network from getting stuck in local minima. The approach is further extended with Temporal Parameter Search (TPS) and Hybrid Network Search (HNS) to refine neuron dynamics and balance efficiency.
LitE-SNN~\cite{liu2024lite} adds spatial and temporal compression to SpikeDHS, and Yan et al.~\cite{yan2024efficient} propose a branchless supernet, ESNN, that encodes architectures with continuous parameters for gradient-based multi-objective optimization.
MA-DARTS~\cite{man2024differentiable}, a prime example of Gradient-Based SNNaS, uses bi-level optimization to simultaneously update architecture parameters (\(\alpha\)) on validation data and network weights (w) on training data. It integrates a multi-dimensional attention mechanism into the node concatenation step, enabling the model to prioritize critical temporal, channel, and spatial information while ignoring less relevant data. This addition significantly improves classification accuracy with only a negligible increase in network size and complexity.
Collectively, these approaches have achieved state-of-the-art performance across various datasets by carefully balancing the high computational demands of differentiable searches with specialized SNN design considerations.

\subsubsection{Hybrid Methods}
Hybrid approaches blend multiple NAS strategies to combine their benefits, such as robust exploration from one method and efficient exploitation from another, thereby speeding searches and conserving resources. Examples include evolutionary algorithms coupled with DRL~\cite{chen2019renas,maziarz2018evolutionary} and random search paired with local grid search. Although these hybrids can produce more balanced results and handle diverse tasks, they also introduce increased complexity and hyperparameter tuning. Careful design is crucial to avoid overlapping strengths and redundant computations, but when done well, hybrid NAS can outperform single-method solutions.

\subsubsection{Search Space Primitives and Inductive Biases}
Architectural primitives available within an SNNaS search space act as inductive biases that strongly shape the quality and efficiency of discovered SNNs. Common choices include cell-based DAGs with searchable operations~\cite{liu2024lite, abdennadher2025lightsnn, man2024differentiable, putra2024neuronas}, convolutional kernels that balance representational capacity against memory cost~\cite{putra2024neuronas, putra2024spikernel}, and pooling strategies that favor max-pooling to preserve sparsity and binary spike behavior~\cite{masters, na2022autosnn, padovano2024spikeexplorer}. Unlike ANN NAS, many SNNaS frameworks further expose temporal structure through backward or recurrent connections~\cite{kim2022neural, saghand2025monas, vasilache2025evolving}, and searchable neuron models and internal parameters such as thresholds, decay rates, and time constants~\cite{xie2023efficient, che2024spatial, ajay2024despine}.

In spiking transformer models, attention mechanisms have emerged as an additional inductive bias. MA-DARTS~\cite{man2024differentiable} integrates multi-dimensional attention (channel, spatial, and temporal) directly into differentiable search, enabling attention-weighted node aggregation during optimization. Transformer-based approaches such as Auto-Spikformer~\cite{che2023auto} treat multi-head spiking self-attention as a searchable primitive, while STMixer~\cite{deng2024spiking} replaces self-attention with hardware-friendly multi-head token mixing to better integrate temporal and spatial features. By contrast, NAS-TinyML~\cite{chen2024based} applies channel-wise attention only to guide macro-level decisions rather than searching attention structure. These methods highlight attention—particularly temporal attention—as an emerging bias for modeling spike-timing variability within SNN search spaces.

Additional inductive biases are imposed through biologically inspired modularity~\cite{pan2024brain, pan2023multi}, explicit efficiency objectives based on spike counts or synaptic operations~\cite{na2022autosnn, xie2025spear, saghand2025monas, song2025evolutionary}, and hardware-aware constraints that prune unsupported operations~\cite{liu2024unleashing, putra2024neuronas}. Many frameworks further bias search using training-free proxies that favor diverse initial activations~\cite{saghand2025monas, kim2022neural, li2024spiking}, layer-wise alignment of neuron dynamics with temporal dependency requirements~\cite{saghand2025monas, ajay2024despine}, or spatial embedding of neurons that constrains connectivity based on physical or biological distance~\cite{pan2023multi, vasilache2025evolving, shen2023brain}.

\subsection{Non-differentiable Hardware Constraints} 
All of the search strategies mentioned above can be adapted to HW-NAS, which aims to optimize neural architectures to meet hardware-specific constraints such as memory, latency, and energy requirements. However, real-world hardware metrics such as bus widths, memory sizes, or specialized neuromorphic parameters are often discrete, making them non-differentiable. Although HW-NAS typically relies on differentiable loss functions for efficient optimization \cite{benmeziane2021comprehensive}, these discrete choices complicate gradient-based approaches.

To address this, methods like Gumbel-Softmax (used in FBNet~\cite{wu2019fbnet}) introduce controlled noise to effectively “soften” discrete variables, while ProxylessNAS~\cite{proxyless} either approximates the discrete search space with a continuous function or applies the REINFORCE algorithm~\cite{williams1992simple} to handle binarized weights. These strategies facilitate backpropagation by approximating non-differentiable hardware constraints, allowing over-parameterized yet gradient-friendly models to incorporate crucial hardware considerations.

\subsection{Performance Evaluation Metrics}\label{performance-evaluation} 
Performance evaluation metrics are essential in SNNaS, as they guide the design of architectures that balance accuracy, latency, energy consumption, memory footprint, and area. While floating point operations (FLOPs) are commonly used, they can be misleading since equal FLOPs may yield varying latencies on different hardware~\cite{bouzidi2021performance, zhang2020fast, wang2020hat}. Accuracy-focused but energy-aware NAS methods, such as Multi-Objective Neural Architectural Search (MONAS)~\cite{hsu2018monas} and NetAdapt~\cite{yang2018netadapt}, incorporate power consumption, and circuit area minimization has also been studied to reduce static power use~\cite{yang2020co}. For edge applications, limited memory budgets are critical, and latency or profiled energy often provides more reliable indicators than FLOPs alone.

Some SNNaS methods track spike-specific metrics such as the number of spikes (NoS), synaptic operations (SynOps), and Bit-SynOps as proxies for energy consumption. Spike count (NoS) is used by AutoSNN~\cite{na2022autosnn} and NeuEvo~\cite{shen2023brain} as a proxy for power, as low spike counts reduce communication overhead; however, this metric does not account for the computational cost of the synapses themselves. ESNN~\cite{yan2024efficient} and LitE-SNN~\cite{liu2024lite} argue that synaptic operations (SynOps), representing AC operations, are more comprehensive indicators of memory and computational resource demands. Structured Pruning for Spiking Neural Networks via Synaptic Operation Estimation and Reinforcement Learning (SPEAR)~\cite{xie2025spear} found that after fine-tuning, SynOps change irregularly, suggesting that using pre-training SynOps as a hard constraint in NAS may result in final models that violate those constraints. Additionally, LitE-SNN~\cite{liu2024lite} and GAQ-SNN~\cite{gaq} introduce mixed-precision quantization to NAS, arguing that hardware performance is determined by the ``bit-width per operation" rather than just the number of operations. Balancing these hardware metrics with task performance ensures robust solutions for diverse scenarios.

To estimate hardware cost, NAS methods use analytical formulas for quick approximations, albeit with less precision on complex designs; LUTs leverage pre-measured data (e.g., layer-specific latency and energy in \cite{wang2020apq}); and simulation-based approaches, such as mobile NAS (MNAS)~\cite{tan2019mnasnet}, offer detail at the expense of time. Machine learning models predict hardware behavior from architecture-hardware mappings~\cite{hsu2018monas, sood2019neunets, yang2020co, proxyless}, but rely on high-quality data. Hybrid approaches combine these methods. For instance, fast analytical or LUT-driven screening may precede more rigorous simulations or ML estimates for top-performing candidates.

While NAS benchmarks are well established for ANNs, SpikeNAS-Bench~\cite{sun2025spikenas} represents the first benchmark explicitly designed for SNNaS. It provides a standardized, cell-based search space with spiking neuron operations and explicit temporal execution. Beyond accuracy, it adopts multi-objective evaluation metrics, including spike activity and synaptic operations, and supports proxy-based evaluation to reduce the cost of exhaustive SNN training. This benchmark enables controlled comparisons of NAS algorithms under consistent SNN-specific assumptions, addressing a key gap in prior SNNaS evaluation.

\subsection{Search Acceleration Strategies} \label{sec:acceleration}
Search acceleration strategies in SNNaS aim to mitigate the prohibitive computational costs of training every candidate architecture from scratch~\cite{zoph2018learning,real2019regularized,real2017large} through approximations, learned parameter reuse, surrogate modeling, and hardware-aware pruning.

\subsubsection{Training-Fidelity Reduction Strategies}
\begin{figure}[t!]
\vspace{-15pt}
		\centering
		\includegraphics[width=\linewidth]{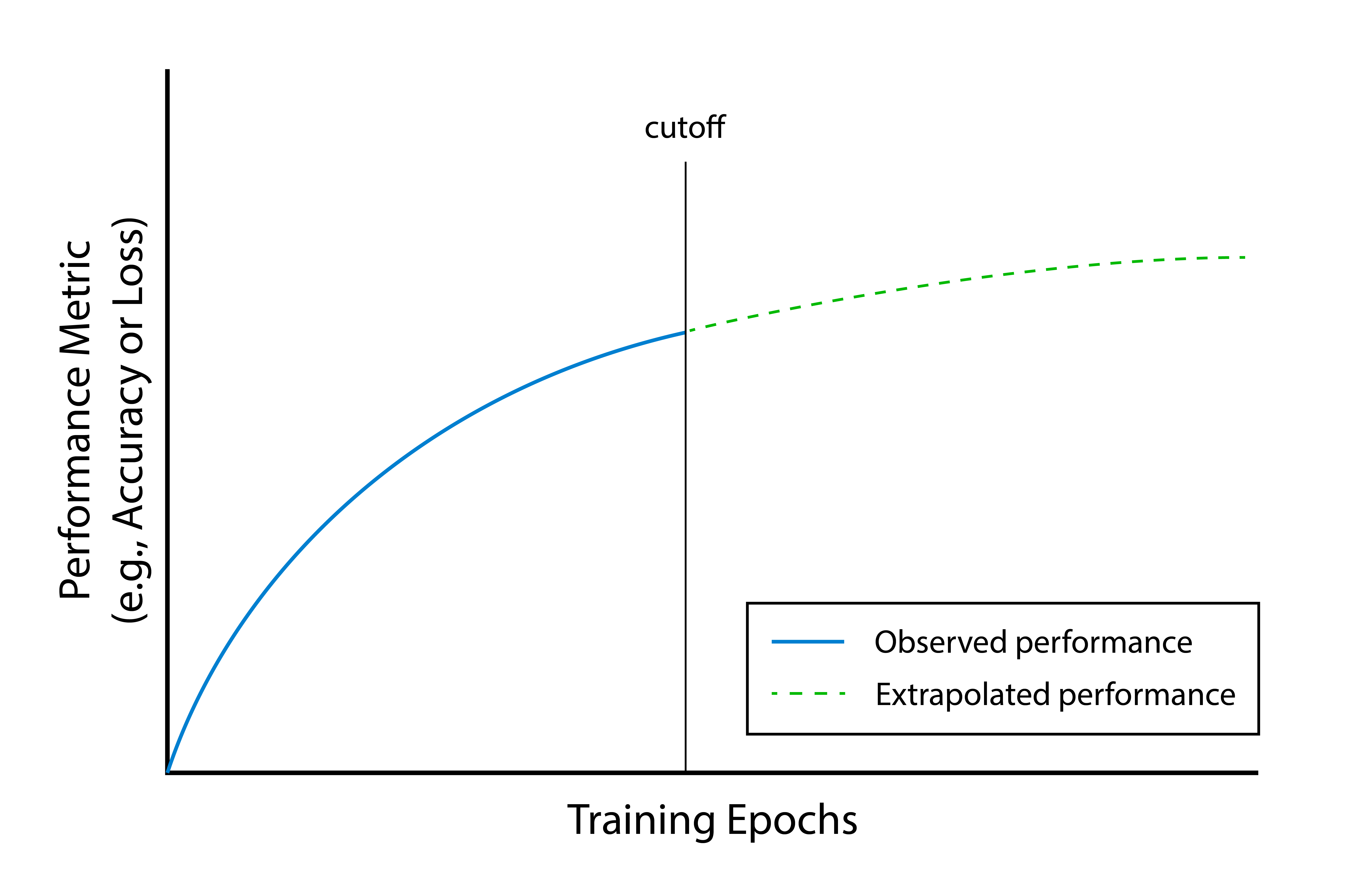}
		\vspace{-10pt}
		\caption{Learning curve extrapolation illustrating the use of partial training data to predict final performance. The observed curve (solid blue) represents the model's accuracy during early epochs, while the extrapolated curve (dashed green) predicts performance for the remaining training epochs. This method enables early termination of unpromising models, saving computational resources and time.}	\label{fig:extrapolation}
\end{figure}

A common acceleration strategy evaluates architectures under reduced training fidelity, e.g., fewer epochs, truncated time steps, smaller datasets, or surrogate objectives. For example, learning curve extrapolation (LCE)~\cite{baker2017accelerating} (illustrated in Figure \ref{fig:extrapolation}) reduces computational costs by extrapolating final performance from partially trained models, avoiding full convergence during architecture evaluation. Progressive strategies accelerate search by incrementally growing architectures and only further training promising partial networks, reducing exploration cost but risking premature pruning.

In SNNaS, Nijenhuis~\cite{masters} achieved a 45$\times$ speed-up by jointly reducing dataset size and temporal resolution. While this approach significantly reduces search time, the approximations risk some loss in accuracy~\cite{liu2018darts,pham2018efficient,real2019regularized,falkner2018bohb,li2017hyperband,runge2018learning,zela2018towards,zoph2018learning,zhou2020econas, man2024differentiable}. Temporal truncation in SNNs can suppress delayed or rate-based spike patterns, leading to systematic underestimation of certain architectures. Unlike ANNs, fidelity reduction does not correlate smoothly with full-training performance.

The covariance matrix adaptation evolution strategy (CMA-ES)~\cite{zhou2020surrogate}, a surrogate-assisted model, mitigates this by using Gaussian processes to provide reliable fitness estimates even with reduced real-network evaluations, lowering the search cost while preserving ranking fidelity.

\subsubsection{Weight Reuse and Supernet-Based Acceleration}
Weight reuse strategies accelerate convergence by avoiding training from scratch. HotNAS~\cite{jiang2020standing}, a hot start method, initializes candidate models from a strong base model to speed early training. Network morphism and weight inheritance further reduce overhead by making incremental architectural changes while retaining trained weights \cite{masters}.

More aggressive reuse occurs in one-shot or weight-sharing NAS, where a supernet is trained once, and candidate subnets inherit shared parameters. Prominent examples include Efficient NAS (ENAS)~\cite{pham2018efficient}, DARTS~\cite{liu2018darts}, and SNAS~\cite{xie2018snas}, with SNN adaptations such as AutoSNN~\cite{na2022autosnn}, ANCoEF~\cite{zhang2024ancoef}, Auto-Spikformer~\cite{che2023auto}, ESNN~\cite{yan2024efficient}, and MA-DARTS~\cite{man2024differentiable}. ESNN~\cite{yan2024efficient}, a one-shot NAS method, reduces search time by up to \(67\times\) by using a branchless spiking supernet that eliminates the need for multiple candidate branches, lowering GPU memory consumption. While effective at reducing training costs, these methods are prone to ranking disorder and performance collapse, as shared weights often fail to reflect the true performance of independently trained SNNs, particularly under temporal coding schemes like time-to-first-spike (TTFS)~\cite{pan2024brain, yan2024efficient, abdennadher2025lightsnn}.

\subsubsection{Initialization-Time and Zero-Shot Performance Proxies}
Zero-shot approaches in the form of zero-cost (ZC) proxies rank architectures without gradient-based training by exploiting initialization-time statistics or analytically derived signals, such as activation overlap~\cite{mellor2021neural, chen2021neural, xu2021knas, sun2021mae, lopes2021epe} or gradient-based measures~\cite{liu2024unleashing}. They are well established in ANN-based NAS, but standard ANN proxies fail in the SNN domain because they rely on gradients that are inaccurate due to the non-differentiability of the spiking function \cite{wang2024autost}.

SNN-specific proxies are explicitly designed to overcome these issues. For instance, SNASNet introduced Sparsity-Aware Hamming Distance (SAHD)~\cite{kim2022neural} to normalize the batch-to-batch sparsity fluctuations in spiking neurons that skew distance-based architecture evaluations, enabling initialization-time, training-free (no weight optimization) scoring that fairly compares architectures under sparse spiking dynamics.
Later, MONAS-ESNN proposed Adjusted SAHD (ASAHD)~\cite{saghand2025monas} to account for neuron-level sparsity variability in Parametric LIF (PLIF) neurons. While Liu et al.~\cite{liu2024unleashing} estimate Loihi-based architectures using gradient-based Gram matrices (GMG), SpikeNAS~\cite{putra2025spikenas} and NeuroNAS~\cite{putra2024neuronas} use similar SNN-aware proxies to achieve orders-of-magnitude faster searches. For spiking transformers, AutoST~\cite{wang2024autost} employs FLOPs as a gradient-independent proxy. While FLOPs are a poor global indicator of SNN efficiency, AutoST shows that within narrow architectural bands—where neuron models, temporal encoding schemes, and execution paradigms are fixed—FLOPs can act as a coarse discriminator between closely related candidate architectures. In these constrained regimes, FLOPs outperform spike-based proxies that are unstable under sparse, non-differentiable dynamics.

The MSE-NAS framework~\cite{pan2023multi} introduces brain-inspired indirect evaluation (BIE) as a specialized "indirect evaluation" function. It accelerates the search by \(60\times\) compared to direct training while ensuring transferability and robustness across datasets.

\subsubsection{Learned Performance Predictors and Few-Shot Methods}

Bridging the gap between zero-shot speed and full-training reliability, several methods employ learned performance predictors trained on small subsets of evaluated architectures. EB-NAS~\cite{pan2024brain} has an online predictor, CART (Classification and Regression Trees), that is updated dynamically during the search process to improve its accuracy as new architecture samples are trained, while PESNN~\cite{wang2025predictor} uses a Random Forest-based few-shot predictor to estimate accuracy at a reduced evaluation cost.

However, SNN predictors trained on limited samples struggle due to the highly non-linear temporal dynamics of SNNs, where minor architectural changes can induce abrupt changes in spike patterns. As a result, these predictors require costly seeding data and generalize poorly beyond the sampled design space, limiting their effectiveness in large-scale or hardware-constrained SNN searches~\cite{abdennadher2025lightsnn}. The SPEAR framework~\cite{xie2025spear} uses a few-shot predictor that employs linear regression to learn the relationship between pre-finetuning and post-finetuning synaptic operations, which solves a common problem of irregular SynOps changes during training.

Auxiliary learning mechanisms can also serve as acceleration tools. MA-DARTS~\cite{man2024differentiable} incorporates an auxiliary classifier to improve gradient propagation and stabilize training during differentiable search.

\subsubsection{Hardware-Aware and System-Level Acceleration}

Frameworks like SPEAR~\cite{xie2025spear} and NeuroNAS~\cite{putra2024neuronas} accelerate searches by incorporating hardware constraints (e.g., memory, area, latency) directly into the search algorithm to prune the search space of unfeasible architectures before they are ever evaluated for accuracy.

The search for hardware-deployable SNNs is often limited by the speed of hardware evaluation. ANCoEF~\cite{zhang2024ancoef} accelerates the search loop by replacing standard simulators (like CanMore) with TrueAsync. TrueAsync uses Akka actor models to provide a \(2\times\) to \(15.8\times\) runtime speedup, directly addressing the high communication latency at the hardware/software boundary.

\subsubsection{Acceleration in Specialized SNN Paradigms (LSMs, TTFS)}

In Liquid State Machine (LSM) NAS, CMA-ES~\cite{zhou2020surrogate} combined with a Gaussian Process (GP)-assisted surrogate significantly reduces the number of required evaluations.
To mitigate supernet instability under TTFS constraints, TNAS-ER~\cite{liu2025efficient} pairs the spiking model with an identity-mapped ANN counterpart. While effective for stabilizing search, this strategy necessarily constrains the architecture to ANN-compatible representations and limits the exploitation of fully spiking temporal dynamics.

\subsubsection{Reliability Limits and Failure Modes of Acceleration in SNNaS}

Despite their effectiveness, acceleration strategies in SNNaS face fundamental reliability constraints. Many zero-cost proxies depend on gradients or activation statistics that are ill-defined under sparse, non-differentiable spiking dynamics, leading to noisy or misleading rankings. Initialization sensitivity is especially severe for adaptive neurons such as PLIF, where sparsity evolves over time.

Weight-sharing methods often exacerbate ranking disorder, as shared weights fail to reflect the true performance of independently trained SNNs, especially under temporal coding schemes such as TTFS~\cite{liu2025efficient}. Similarly, predictor-based methods struggle to extrapolate beyond limited training samples due to the discontinuous nature of spike dynamics. As a result, acceleration methods transferred from ANN-NAS may bias the search toward architectures that score well on static or early-time proxies but underperform when full temporal dynamics and training convergence are considered~\cite{liu2025efficient}. 

\subsection{Temporal Considerations} \label{temporal}
Temporal dynamics are central to SNNs’ unique capabilities, yet many SNNaS approaches adapt ANN-centric methods that overlook time-related factors. Recent work has begun to address this gap. For instance, LitE-SNN~\cite{liu2024lite} integrates both spatial and temporal compression, and MixedSNN~\cite{xie2023efficient} leverages Period-Based Spike Evaluation (PBSE) to capture critical temporal attributes across the entire simulation duration. Che et al.~\cite{che2024spatial} further introduce a spatial-temporal search methodology that uses TPS to fine-tune different neurons’ time constants, optimizing both architecture and temporal behaviors. 
MA-DARTS~\cite{man2024differentiable} incorporates temporal attention alongside spatial and channel dimensions, allowing the network to adaptively weight different time steps to focus on temporally important information while maintaining computational efficiency. By explicitly modeling temporal dynamics via attention mechanisms, MA-DARTS achieves significant improvements, suggesting that future SNNaS methods should integrate temporal processing into the architecture search process to support a more holistic SNN design paradigm.

\begin{table*}[ht]
\centering
\caption{Global View of SNNaS Methods Grouped by Search Strategy. Search spaces, acceleration techniques, and metrics reflect dominant patterns within each category.}
\label{table:SNNAS-papers}
\begin{tabular}{p{1.6cm} p{5.2cm} p{2.2cm} p{2.4cm} p{2.6cm}}
\toprule
\textbf{Strategy} & \textbf{Representative Works} & \textbf{Search Spaces} & \textbf{Acceleration} & \textbf{Key Metrics} \\
\midrule
Evolutionary 
& EONS~\cite{schuman2016evolutionary}, MOO-SNNs~\cite{dimovska2019multi}, GP-Assisted CMA-ES~\cite{zhou2020surrogate}, Spiking WANN~\cite{anwar2021evolving}, AutoSNN~\cite{na2022autosnn}, GAQ~\cite{gaq}, Auto-Spikformer~\cite{che2023auto}, NeuEvo~\cite{shen2023brain}, MixedSNN~\cite{xie2023efficient}, ANAS~\cite{zhang2023anas}, ESL-SNNs~\cite{shen2023esl}, SPENSER~\cite{branquinho2023spenser}, EB-NAS~\cite{pan2024brain}, MSE-NAS~\cite{pan2023multi}, DESPINE~\cite{ajay2024despine}, AutoST~\cite{wang2024autost}, SSTNAS~\cite{li2024spiking}, PESNN~\cite{wang2025predictor}, TNAS-ER~\cite{liu2025efficient}, EMO-SNAS~\cite{song2025evolutionary}, MONAS-ESNN~\cite{saghand2025monas}, Membrane-NAS~\cite{liu2025auto}, ABC-Arc~\cite{zhang2025spike}, BCNAS-SNN~\cite{dengproper}, Vasilache et al.~\cite{vasilache2025evolving}, NAS-TinyML~\cite{chen2024based}
& Global, Hierarchical, Sequential, Cell-based 
& Training-free, One-shot, Surrogate models, Weight inheritance, Lower fidelity 
& Accuracy, NoS, Energy, Network size \\
\midrule
Gradient-Based 
& Nijenhuis~\cite{masters}, SpikeDHS~\cite{che2022differentiable}, SpikeDHS+TPS/HNS~\cite{che2024spatial}, ESNN~\cite{yan2024efficient}, LitE-SNN~\cite{liu2024lite}, MA-DARTS~\cite{man2024differentiable}, ST-DANO~\cite{zhangspatio}, SpikeX~\cite{xu2025spikex}
& Cell-based, Hierarchical, Global 
& One-shot, Weight sharing, Lower fidelity 
& Accuracy, SynOps, Task loss, EDP \\
\midrule
Bayesian Opt. 
& deepHyper~\cite{yanguas2022automl}, Benmeziane et al.~\cite{benmeziane2023skip}, SpikeExplorer~\cite{padovano2024spikeexplorer}, ASNPC~\cite{wang2025asnpc}
& Global, Cell-based 
& Surrogate models, Lower fidelity, Weight inheritance 
& Accuracy, Area, Latency, Power \\
\midrule
RL-Based 
& ANCoEF~\cite{zhang2024ancoef}, SPEAR~\cite{xie2025spear}
& Sequential 
& One-shot, Lower fidelity 
& Accuracy, EDP, SynOps, Area \\
\midrule
Random/Grid 
& SNASNet~\cite{kim2022neural}, Liu \& Yi~\cite{liu2024unleashing}, SpiKernel~\cite{putra2024methodology}, NeuroNAS~\cite{putra2024neuronas}, SpikeNAS~\cite{putra2025spikenas}, STMixer~\cite{deng2024spiking}, LightSNN~\cite{abdennadher2025lightsnn}
& Cell-based, Hierarchical, Sequential 
& Training-free NAS, Lower fidelity, Progressive search 
& SAHD, GMG, Accuracy, FLOPs \\
\bottomrule
\end{tabular}
\end{table*}

\section{Challenges and Future Directions}\label{sec:future} 
Table \ref{table:SNNAS-papers} provides a global view of representative SNNaS studies surveyed in this work, highlighting their search spaces, search algorithms, acceleration strategies, and performance criteria. While these approaches have driven notable progress in the field, SNNaS continues to face fundamental challenges in search space design, hardware alignment, interpretability, and reproducibility.

\subsection{Open Challenges in SNNaS}

Many SNNaS methods remain misaligned with the realities of neuromorphic execution. Search objectives are typically evaluated on synchronous simulators and simplified energy proxies, overlooking asynchronous timing errors, communication effects, and non-numerical hardware design dimensions such as routing or on-chip dataflow. As a result, architectures that appear optimal in simulation may perform poorly when deployed. Moreover, understanding \emph{why} certain automatically discovered architectures perform well remains elusive, underscoring a need for deeper theoretical insights into SNN behaviors \cite{he2021automl}.

Reproducibility represents a structural barrier to cumulative progress in SNNaS. Across NAS more broadly, unmaintained codebases, dependency decay, and inadequate documentation have led to a reproducibility crisis, with approximately 75\% of published NAS methods failing to run out of the box \cite{towers2022long}. SNNaS inherits these issues, which divert research effort toward reimplementation rather than extension, slowing cumulative advances.

These problems are exacerbated by the absence of large, diverse, and hardware-grounded SNNaS benchmarks, which continues to confine SNNaS to ANN-centric tasks, search spaces, and metrics. Although acceleration strategies such as one-shot NAS, weight sharing, and zero-cost proxies reduce computational burden \cite{elsken2019neural,white2023neural}, they introduce additional biases: parameter sharing may degrade final model quality or misrank promising designs~\cite{li2020block, yu2019evaluating, zhang2020overcoming, cai2020once}.

Without stable implementations, standardized benchmarks, and environment-robust releases, the empirical validity of SNNaS comparisons remains fragile, particularly for hardware-aware and acceleration-heavy methods.

\subsection{Emerging Research Directions}

Joint optimization of architectures and hyperparameters could lessen human bias and yield more robust solutions, especially under few-shot learning or meta-learning conditions \cite{he2021automl,liu2020labels}. There is also growing interest in architectures resilient to adversarial attacks \cite{elsken2019neural}, multi-task learning, and multi-objective optimization that accounts for both performance and resource efficiency \cite{elsken2019neural}. 

In parallel, hardware awareness in SNNaS remains limited. Many methods implicitly assume von Neumann execution models, overlooking memory bottlenecks and the constraints of neuromorphic systems \cite{benmeziane2021comprehensive}. Co-optimizing SNN architectures with emerging hardware-oriented paradigms such as in-memory computing, quantization, and pruning could enable more efficient and biologically plausible solutions \cite{jiang2020device, benmeziane2021comprehensive, aliyev2025exploring}. Many existing SNNaS methods target a single “universal” architecture rather than exploring device-specific designs, highlighting the need for hardware-targeted or co-explored SNNaS frameworks \cite{benmeziane2021comprehensive}.

Sustained progress in SNNaS also depends on treating reproducibility as a first-class research objective. Beyond open-source releases, future work should prioritize environment-stable implementations (e.g., containerized methods), clear documentation and pseudocode, explicit seed control and logging, and framework-level standardization to enable fair comparison and long-term reuse \cite{towers2022long}. 

Finally, advancing SNNaS benchmarks to include diverse neuromorphic tasks, richer temporal dynamics, and hardware-grounded evaluation is critical. Such benchmarks should move beyond ANN-centric protocols to include SNN-native search spaces with explicit neuron-level parameterization, multi-objective metrics that capture accuracy, sparsity, and energy-relevant costs (e.g., SynOps), and standardized validation of acceleration proxies through rank correlation with fully trained models.

\section{Summary of Key Insights} \label{sec:insights}

This survey has examined the rapidly evolving field of Spiking Neural Network Architecture Search (SNNaS) through the lens of hardware-software co-design. Our analysis reveals that successful SNNaS fundamentally differs from traditional ANN-based NAS due to the unique computational characteristics of spiking networks. The non-differentiable spike function, temporal dynamics spanning multiple timesteps, and event-driven processing paradigm collectively necessitate specialized search spaces, evaluation metrics, and optimization strategies (Section \ref{sec:background}). A recurring theme across the surveyed literature is that approaches that simply adapt ANN architectures, e.g., replacing ReLU with LIF neurons or borrowing backbone structures like VGG and ResNet, consistently outperform methods designed specifically for spike-based computation (Section \ref{sec:nas_challenges}). The evolution from global and sequential search spaces toward cell-based, hierarchical, and biologically-inspired representations reflects the field's growing recognition that SNN-specific inductive biases are essential for discovering high-performing architectures (Section \ref{sec:search_space}).

The co-design perspective emerges as perhaps the most critical insight from this survey. Hardware constraints in neuromorphic computing are not merely optimization targets to be addressed post-hoc; they fundamentally shape which architectures are viable and efficient (Section \ref{sec:hw_nas}). Asynchronous, event-driven platforms like Intel's Loihi achieve their power efficiency by eliminating global clocks, but this design choice renders certain operations problematic. For instance, spiking matrix multiplication and max-pooling layers become vulnerable to timing imprecision (Section \ref{sec:hw_sw_boundary}). Co-exploration frameworks that jointly search neural architectures and hardware configurations (e.g., ANAS, ANCoEF) consistently achieve superior energy-delay products compared to methods that optimize these dimensions separately (Section \ref{sec:coexploration}). Furthermore, the choice of performance metrics profoundly impacts search outcomes. While spike count (NoS) serves as a simple proxy for power consumption, synaptic operations (SynOps) provide more comprehensive estimates of computational cost, and Bit-SynOps account for the impact of quantization on hardware efficiency (Section \ref{performance-evaluation}).

Finally, the field has made remarkable progress in addressing the computational burden of architecture search through training-free and low-fidelity evaluation strategies (Section \ref{sec:acceleration}). Zero-shot metrics like SAHD and GMG enable architecture ranking without any training, while one-shot supernet approaches and surrogate predictors dramatically reduce the cost of evaluating candidate architectures. These acceleration techniques have reduced search times from thousands of GPU-hours to practical levels, democratizing SNNaS research. Additionally, the explicit modeling of temporal dynamics, through temporal attention mechanisms, timestep optimization, and spike-timing-aware evaluation, represents a critical frontier that remains underexploited in current methods (Section \ref{temporal}). As summarized in Table \ref{table:SNNAS-papers}, the diversity of search strategies (evolutionary, gradient-based, Bayesian, and reinforcement learning approaches) reflects the field's ongoing exploration of how best to navigate the complex, multi-objective optimization landscape inherent to SNNaS (Section \ref{sec:search_strategies}). Looking forward, significant challenges remain in interpretability, generalization beyond image classification, and scalability (Section \ref{sec:future}). As neuromorphic hardware continues to mature and standardized tools emerge, we anticipate that SNNaS will play an increasingly central role in unlocking the full potential of brain-inspired computing for energy-efficient, real-time AI applications.

\section{Conclusion} \label{sec:conclusion}
Spiking neural network architecture search (SNNaS) demands a tightly integrated hardware/software co-design strategy, as standard ANN-based NAS methods overlook spiking-specific challenges. Recent progress, encompassing hardware-aware design, explicit temporal considerations, and novel search algorithms such as surrogate gradients, Bayesian optimization, and hybrid techniques, demonstrates the growing sophistication in SNN-focused approaches. However, fundamental issues remain, such as bridging hardware-software integration, reducing computational costs, and improving energy efficiency. Addressing these challenges is essential for scaling SNNs to real-world tasks. Overcoming these challenges and deploying SNNs on dedicated neuromorphic hardware will be crucial to unlocking the full potential of spiking networks and driving neuromorphic computing forward.

\bibliographystyle{ieeetr}
{\small
\bibliography{refs}

@inproceedings{strubell2019carbon,
  title={Energy and policy considerations for deep learning in NLP},
  author={Strubell, Emma and Ganesh, Ananya and McCallum, Andrew},
  booktitle={Proceedings of the 57th annual meeting of the association for computational linguistics},
  pages={3645--3650},
  year={2019}
}

@article{hodgkin1952quantitative,
  title={A quantitative description of membrane current and its application to conduction and excitation in nerve},
  author={Hodgkin, Alan L and Huxley, Andrew F},
  journal={The Journal of physiology},
  volume={117},
  number={4},
  pages={500},
  year={1952},
  publisher={Wiley}
}

@book{koch1998methods,
  title={Methods in neuronal modeling: from ions to networks},
  author={Koch, Christof and Segev, Idan},
  publisher={MIT press},
  year={1998}
}

@article{abbott2000synaptic,
  title={Synaptic plasticity: taming the beast},
  author={Abbott, Larry F and Nelson, Sacha B},
  journal={Nature neuroscience},
  volume={3},
  number={11},
  pages={1178--1183},
  year={2000},
  publisher={Nature Publishing Group}
}

@article{bi1998synaptic,
  title={Synaptic modifications in cultured hippocampal neurons: dependence on spike timing, synaptic strength, and postsynaptic cell type},
  author={Bi, Guo-qiang and Poo, Mu-ming},
  journal={Journal of neuroscience},
  volume={18},
  number={24},
  pages={10464--10472},
  year={1998},
  publisher={Soc Neuroscience}
}

@book{gerstner2002spiking,
  title={Spiking neuron models: Single neurons, populations, plasticity},
  author={Gerstner, Wulfram and Kistler, Werner M},
  year={2002},
  publisher={Cambridge university press}
}

@article{shrestha2019review,
  title={Review of deep learning algorithms and architectures},
  author={Shrestha, Ajay and Mahmood, Ausif},
  journal={IEEE access},
  volume={7},
  pages={53040--53065},
  year={2019},
  publisher={IEEE}
}

@inproceedings{simonyan2015deep,
  title={Very deep convolutional networks for large-scale image recognition},
  author={Simonyan, K and Zisserman, A},
  booktitle={3rd International Conference on Learning Representations (ICLR 2015)},
  year={2015},
  organization={Computational and Biological Learning Society}
}

@inproceedings{he2016deep,
  title={Deep residual learning for image recognition},
  author={He, Kaiming and Zhang, Xiangyu and Ren, Shaoqing and Sun, Jian},
  booktitle={Proceedings of the IEEE conference on computer vision and pattern recognition},
  pages={770--778},
  year={2016}
}

@article{lin2020mcunet,
  title={Mcunet: Tiny deep learning on iot devices},
  author={Lin, Ji and Chen, Wei-Ming and Lin, Yujun and Gan, Chuang and Han, Song and others},
  journal={Advances in Neural Information Processing Systems},
  volume={33},
  pages={11711--11722},
  year={2020}
}

@inproceedings{wang2020hat,
  title={HAT: Hardware-Aware Transformers for Efficient Natural Language Processing},
  author={Wang, Hanrui and Wu, Zhanghao and Liu, Zhijian and Cai, Han and Zhu, Ligeng and Gan, Chuang and Han, Song},
  year={2020},
  organization={Association for Computational Linguistics (ACL)}
}

@inproceedings{cai2020once,
  title={Once for All: Train One Network and Specialize it for Efficient Deployment},
  author={Cai, Han and Gan, Chuang and Wang, Tianzhe and Zhang, Zhekai and Han, Song},
  booktitle={International Conference on Learning Representations},
  year={2020}
}

@inproceedings{bouzidi2021performance,
  title={Performance prediction for convolutional neural networks on edge gpus},
  author={Bouzidi, Halima and Ouarnoughi, Hamza and Niar, Smail and Cadi, Abdessamad Ait El},
  booktitle={Proceedings of the 18th ACM International Conference on Computing Frontiers},
  pages={54--62},
  year={2021}
}

@article{he2021automl,
  title={AutoML: A survey of the state-of-the-art},
  author={He, Xin and Zhao, Kaiyong and Chu, Xiaowen},
  journal={Knowledge-based systems},
  volume={212},
  pages={106622},
  year={2021},
  publisher={Elsevier}
}

@article{liu2021survey,
  title={A survey on evolutionary neural architecture search},
  author={Liu, Yuqiao and Sun, Yanan and Xue, Bing and Zhang, Mengjie and Yen, Gary G and Tan, Kay Chen},
  journal={IEEE transactions on neural networks and learning systems},
  year={2021},
  publisher={IEEE}
}

@article{ren2021comprehensive,
  title={A comprehensive survey of neural architecture search: Challenges and solutions},
  author={Ren, Pengzhen and Xiao, Yun and Chang, Xiaojun and Huang, Po-Yao and Li, Zhihui and Chen, Xiaojiang and Wang, Xin},
  journal={ACM Computing Surveys (CSUR)},
  volume={54},
  number={4},
  pages={1--34},
  year={2021},
  publisher={ACM New York, NY, USA}
}

@article{elsken2019neural,
  title={Neural architecture search: A survey},
  author={Elsken, Thomas and Metzen, Jan Hendrik and Hutter, Frank},
  journal={The Journal of Machine Learning Research},
  volume={20},
  number={1},
  pages={1997--2017},
  year={2019},
  publisher={JMLR. org}
}

@article{white2023neural,
  title={Neural architecture search: Insights from 1000 papers},
  author={White, Colin and Safari, Mahmoud and Sukthanker, Rhea and Ru, Binxin and Elsken, Thomas and Zela, Arber and Dey, Debadeepta and Hutter, Frank},
  journal={arXiv preprint arXiv:2301.08727},
  year={2023}
}

@article{benmeziane2021comprehensive,
  title={A comprehensive survey on hardware-aware neural architecture search},
  author={Benmeziane, Hadjer and Maghraoui, Kaoutar El and Ouarnoughi, Hamza and Niar, Smail and Wistuba, Martin and Wang, Naigang},
  journal={arXiv preprint arXiv:2101.09336},
  year={2021}
}

@inproceedings{radosavovic2020designing,
  title={Designing network design spaces},
  author={Radosavovic, Ilija and Kosaraju, Raj Prateek and Girshick, Ross and He, Kaiming and Doll{\'a}r, Piotr},
  booktitle={Proceedings of the IEEE/CVF conference on computer vision and pattern recognition},
  pages={10428--10436},
  year={2020}
}

@inproceedings{marchisio2020nascaps,
  title={NASCaps: A framework for neural architecture search to optimize the accuracy and hardware efficiency of convolutional capsule networks},
  author={Marchisio, Alberto and Massa, Andrea and Mrazek, Vojtech and Bussolino, Beatrice and Martina, Maurizio and Shafique, Muhammad},
  booktitle={Proceedings of the 39th International Conference on Computer-Aided Design},
  pages={1--9},
  year={2020}
}

@inproceedings{zhang2023anas,
  title={ANAS: Asynchronous Neuromorphic Hardware Architecture Search Based on a System-Level Simulator},
  author={Zhang, Jian and Zhang, Jilin and Huo, Dexuan and Chen, Hong},
  booktitle={2023 60th ACM/IEEE Design Automation Conference (DAC)},
  pages={1--6},
  year={2023},
  organization={IEEE}
}

@inproceedings{chen2019renas,
  title={Renas: Reinforced evolutionary neural architecture search},
  author={Chen, Yukang and Meng, Gaofeng and Zhang, Qian and Xiang, Shiming and Huang, Chang and Mu, Lisen and Wang, Xinggang},
  booktitle={Proceedings of the IEEE/CVF conference on computer vision and pattern recognition},
  pages={4787--4796},
  year={2019}
}

@inproceedings{tan2019mnasnet,
  title={Mnasnet: Platform-aware neural architecture search for mobile},
  author={Tan, Mingxing and Chen, Bo and Pang, Ruoming and Vasudevan, Vijay and Sandler, Mark and Howard, Andrew and Le, Quoc V},
  booktitle={Proceedings of the IEEE/CVF conference on computer vision and pattern recognition},
  pages={2820--2828},
  year={2019}
}

@inproceedings{wu2019fbnet,
  title={Fbnet: Hardware-aware efficient convnet design via differentiable neural architecture search},
  author={Wu, Bichen and Dai, Xiaoliang and Zhang, Peizhao and Wang, Yanghan and Sun, Fei and Wu, Yiming and Tian, Yuandong and Vajda, Peter and Jia, Yangqing and Keutzer, Kurt},
  booktitle={Proceedings of the IEEE/CVF conference on computer vision and pattern recognition},
  pages={10734--10742},
  year={2019}
}

@inproceedings{wang2020apq,
  title={Apq: Joint search for network architecture, pruning and quantization policy},
  author={Wang, Tianzhe and Wang, Kuan and Cai, Han and Lin, Ji and Liu, Zhijian and Wang, Hanrui and Lin, Yujun and Han, Song},
  booktitle={Proceedings of the IEEE/CVF Conference on Computer Vision and Pattern Recognition},
  pages={2078--2087},
  year={2020}
}

@inproceedings{yang2018netadapt,
  title={Netadapt: Platform-aware neural network adaptation for mobile applications},
  author={Yang, Tien-Ju and Howard, Andrew and Chen, Bo and Zhang, Xiao and Go, Alec and Sandler, Mark and Sze, Vivienne and Adam, Hartwig},
  booktitle={Proceedings of the European Conference on Computer Vision (ECCV)},
  pages={285--300},
  year={2018}
}

@inproceedings{lopes2021epe,
  title={Epe-nas: Efficient performance estimation without training for neural architecture search},
  author={Lopes, Vasco and Alirezazadeh, Saeid and Alexandre, Lu{\'\i}s A},
  booktitle={International conference on artificial neural networks},
  pages={552--563},
  year={2021},
  organization={Springer}
}

@inproceedings{sun2021mae,
  title={MAE-DET: Revisiting Maximum Entropy Principle in Zero-Shot NAS for Efficient Object Detection},
  author={Sun, Zhenhong and Lin, Ming and Sun, Xiuyu and Tan, Zhiyu and Li, Hao and Jin, Rong},
  booktitle={International Conference on Machine Learning},
  pages={20810--20826},
  year={2022},
  organization={PMLR}
}

@inproceedings{xu2021knas,
  title={KNAS: green neural architecture search},
  author={Xu, Jingjing and Zhao, Liang and Lin, Junyang and Gao, Rundong and Sun, Xu and Yang, Hongxia},
  booktitle={International Conference on Machine Learning},
  pages={11613--11625},
  year={2021},
  organization={PMLR}
}

@inproceedings{liu2018darts,
  title={DARTS: Differentiable Architecture Search},
  author={Liu, Hanxiao and Simonyan, Karen and Yang, Yiming},
  booktitle={International Conference on Learning Representations},
  year={2019}
}

@inproceedings{xie2018snas,
  title={SNAS: stochastic neural architecture search},
  author={Xie, Sirui and Zheng, Hehui and Liu, Chunxiao and Lin, Liang},
  booktitle={International Conference on Learning Representations},
  year={2019}
}

@article{binghamineffectiveness,
  title={Ineffectiveness of Gradient-based Neural Architecture Search},
  author={Bingham, Garrett}
}

@inproceedings{li2017hyperband,
  title={Hyperband: Bandit-Based Configuration Evaluation for Hyperparameter Optimization.},
  author={Li, Lisha and Jamieson, Kevin G and DeSalvo, Giulia and Rostamizadeh, Afshin and Talwalkar, Ameet},
  booktitle={ICLR (Poster)},
  pages={53},
  year={2017}
}

@inproceedings{falkner2018bohb,
  title={BOHB: Robust and efficient hyperparameter optimization at scale},
  author={Falkner, Stefan and Klein, Aaron and Hutter, Frank},
  booktitle={International conference on machine learning},
  pages={1437--1446},
  year={2018},
  organization={PMLR}
}

@article{shahriari2015taking,
  title={Taking the human out of the loop: A review of Bayesian optimization},
  author={Shahriari, Bobak and Swersky, Kevin and Wang, Ziyu and Adams, Ryan P and De Freitas, Nando},
  journal={Proceedings of the IEEE},
  volume={104},
  number={1},
  pages={148--175},
  year={2015},
  publisher={IEEE}
}

@inproceedings{zela2018towards,
  title={Towards automated deep learning: Efficient joint neural architecture and hyperparameter search},
  author={Zela, Arber and Klein, Aaron and Falkner, Stefan and Hutter, Frank},
  booktitle={International Conference on Machine Learning (ICML) 2018, AutoML Workshop},
  year={2018}
}

@article{bergstra2012random,
  title={Random search for hyper-parameter optimization.},
  author={Bergstra, James and Bengio, Yoshua},
  journal={Journal of machine learning research},
  volume={13},
  number={2},
  year={2012}
}

@inproceedings{zhou2020econas,
  title={Econas: Finding proxies for economical neural architecture search},
  author={Zhou, Dongzhan and Zhou, Xinchi and Zhang, Wenwei and Loy, Chen Change and Yi, Shuai and Zhang, Xuesen and Ouyang, Wanli},
  booktitle={Proceedings of the IEEE/CVF Conference on computer vision and pattern recognition},
  pages={11396--11404},
  year={2020}
}

@inproceedings{pham2018efficient,
  title={Efficient neural architecture search via parameters sharing},
  author={Pham, Hieu and Guan, Melody and Zoph, Barret and Le, Quoc and Dean, Jeff},
  booktitle={International conference on machine learning},
  pages={4095--4104},
  year={2018},
  organization={PMLR}
}

@inproceedings{zhang2020overcoming,
  title={Overcoming multi-model forgetting in one-shot NAS with diversity maximization},
  author={Zhang, Miao and Li, Huiqi and Pan, Shirui and Chang, Xiaojun and Su, Steven},
  booktitle={Proceedings of the ieee/cvf conference on computer vision and pattern recognition},
  pages={7809--7818},
  year={2020}
}

@inproceedings{zoph2018learning,
  title={Learning transferable architectures for scalable image recognition},
  author={Zoph, Barret and Vasudevan, Vijay and Shlens, Jonathon and Le, Quoc V},
  booktitle={Proceedings of the IEEE conference on computer vision and pattern recognition},
  pages={8697--8710},
  year={2018}
}

@inproceedings{prellberg2018lamarckian,
  title={Lamarckian evolution of convolutional neural networks},
  author={Prellberg, Jonas and Kramer, Oliver},
  booktitle={Parallel Problem Solving from Nature--PPSN XV: 15th International Conference, Coimbra, Portugal, September 8--12, 2018, Proceedings, Part II 15},
  pages={424--435},
  year={2018},
  organization={Springer}
}

@inproceedings{elsken2018efficient,
  title={Efficient Multi-Objective Neural Architecture Search via Lamarckian Evolution},
  author={Thomas Elsken and Jan Hendrik Metzen and Frank Hutter},
  booktitle={International Conference on Learning Representations},
  year={2019}
}

@inproceedings{real2017large,
  title={Large-scale evolution of image classifiers},
  author={Real, Esteban and Moore, Sherry and Selle, Andrew and Saxena, Saurabh and Suematsu, Yutaka Leon and Tan, Jie and Le, Quoc V and Kurakin, Alexey},
  booktitle={International conference on machine learning},
  pages={2902--2911},
  year={2017},
  organization={PMLR}
}

@inproceedings{miller1989designing,
  title={Designing Neural Networks Using Genetic Algorithms.},
  author={Miller, Geoffrey F and Todd, Peter M and Hegde, Shailesh U},
  booktitle={ICGA},
  volume={89},
  pages={379--384},
  year={1989}
}

@inproceedings{real2019regularized,
  title={Regularized evolution for image classifier architecture search},
  author={Real, Esteban and Aggarwal, Alok and Huang, Yanping and Le, Quoc V},
  booktitle={Proceedings of the aaai conference on artificial intelligence},
  volume={33},
  number={01},
  pages={4780--4789},
  year={2019}
}

@inproceedings{shrestha2019optimizing,
  title={Optimizing deep neural network architecture with enhanced genetic algorithm},
  author={Shrestha, Ajay and Mahmood, Ausif},
  booktitle={2019 18th IEEE International Conference On Machine Learning And Applications (ICMLA)},
  pages={1365--1370},
  year={2019},
  organization={IEEE}
}

@inproceedings{young2015optimizing,
  title={Optimizing deep learning hyper-parameters through an evolutionary algorithm},
  author={Young, Steven R and Rose, Derek C and Karnowski, Thomas P and Lim, Seung-Hwan and Patton, Robert M},
  booktitle={Proceedings of the workshop on machine learning in high-performance computing environments},
  pages={1--5},
  year={2015}
}

@inproceedings{baker2016designing,
  title={Designing Neural Network Architectures using Reinforcement Learning},
  author={Baker, Bowen and Gupta, Otkrist and Naik, Nikhil and Raskar, Ramesh},
  booktitle={International Conference on Learning Representations},
  year={2017}
}

@article{stanley2002evolving,
  title={Evolving neural networks through augmenting topologies},
  author={Stanley, Kenneth O and Miikkulainen, Risto},
  journal={Evolutionary computation},
  volume={10},
  number={2},
  pages={99--127},
  year={2002},
  publisher={MIT Press}
}

@inproceedings{liu2018hierarchical,
  title={Hierarchical Representations for Efficient Architecture Search},
  author={Liu, Hanxiao and Simonyan, Karen and Vinyals, Oriol and Fernando, Chrisantha and Kavukcuoglu, Koray},
  booktitle={International Conference on Learning Representations},
  year={2018}
}

@article{guha1988genetic,
  title={Genetic synthesis of neural networks},
  author={Guha, Aloke and Harp, Steven Alex and Samad, Tariq},
  journal={Honeywell Corporate Syst. Development Division, Tech. Rep. CSDD-88-I4852-CC-1},
  year={1988}
}

@article{todd1988evolutionary,
  title={Evolutionary methods for connectionist architectures},
  author={Todd, P},
  journal={Psychology Dept. Stanford University, unpublished Manuscript},
  year={1988}
}

@inproceedings{zoph2017neural,
  title={Neural Architecture Search with Reinforcement Learning},
  author={Zoph, Barret and Le, Quoc},
  booktitle={International Conference on Learning Representations},
  year={2017}
}

@article{maziarz2018evolutionary,
  title={Evolutionary-neural hybrid agents for architecture search},
  author={Maziarz, Krzysztof and Tan, Mingxing and Khorlin, Andrey and Georgiev, Marin and Gesmundo, Andrea},
  journal={arXiv preprint arXiv:1811.09828},
  year={2018}
}

@article{mouret2015illuminating,
  title={Illuminating search spaces by mapping elites},
  author={Mouret, Jean-Baptiste and Clune, Jeff},
  journal={arXiv preprint arXiv:1504.04909},
  year={2015}
}

@article{williams1992simple,
  title={Simple statistical gradient-following algorithms for connectionist reinforcement learning},
  author={Williams, Ronald J},
  journal={Machine learning},
  volume={8},
  pages={229--256},
  year={1992},
  publisher={Springer}
}

@book{sutton2018reinforcement,
  title={Reinforcement learning: An introduction},
  author={Sutton, Richard S and Barto, Andrew G},
  year={2018},
  publisher={MIT press}
}

@inproceedings{mellor2021neural,
  title={Neural architecture search without training},
  author={Mellor, Joe and Turner, Jack and Storkey, Amos and Crowley, Elliot J},
  booktitle={International Conference on Machine Learning},
  pages={7588--7598},
  year={2021},
  organization={PMLR}
}

@inproceedings{yu2019evaluating,
  title={Evaluating The Search Phase of Neural Architecture Search},
  author={Yu, Kaicheng and Sciuto, Christian and Jaggi, Martin and Musat, Claudiu and Salzmann, Mathieu},
  booktitle={International Conference on Learning Representations},
  year={2019}
}

@inproceedings{li2020random,
  title={Random search and reproducibility for neural architecture search},
  author={Li, Liam and Talwalkar, Ameet},
  booktitle={Uncertainty in artificial intelligence},
  pages={367--377},
  year={2020},
  organization={PMLR}
}

@inproceedings{liu2020labels,
  title={Are labels necessary for neural architecture search?},
  author={Liu, Chenxi and Doll{\'a}r, Piotr and He, Kaiming and Girshick, Ross and Yuille, Alan and Xie, Saining},
  booktitle={Computer Vision--ECCV 2020: 16th European Conference, Glasgow, UK, August 23--28, 2020, Proceedings, Part IV 16},
  pages={798--813},
  year={2020},
  organization={Springer}
}

@inproceedings{li2020block,
  title={Block-wisely supervised neural architecture search with knowledge distillation},
  author={Li, Changlin and Peng, Jiefeng and Yuan, Liuchun and Wang, Guangrun and Liang, Xiaodan and Lin, Liang and Chang, Xiaojun},
  booktitle={Proceedings of the IEEE/CVF Conference on Computer Vision and Pattern Recognition},
  pages={1989--1998},
  year={2020}
}

@article{baker2017accelerating,
  title={Accelerating neural architecture search using performance prediction},
  author={Baker, Bowen and Gupta, Otkrist and Raskar, Ramesh and Naik, Nikhil},
  journal={arXiv preprint arXiv:1705.10823},
  year={2017}
}

@inproceedings{yang2020co,
  title={Co-exploration of neural architectures and heterogeneous asic accelerator designs targeting multiple tasks},
  author={Yang, Lei and Yan, Zheyu and Li, Meng and Kwon, Hyoukjun and Lai, Liangzhen and Krishna, Tushar and Chandra, Vikas and Jiang, Weiwen and Shi, Yiyu},
  booktitle={2020 57th ACM/IEEE Design Automation Conference (DAC)},
  pages={1--6},
  year={2020},
  organization={IEEE}
}

@inproceedings{zhang2019neural,
  title={When neural architecture search meets hardware implementation: from hardware awareness to co-design},
  author={Zhang, Xinyi and Jiang, Weiwen and Shi, Yiyu and Hu, Jingtong},
  booktitle={2019 IEEE Computer Society Annual Symposium on VLSI (ISVLSI)},
  pages={25--30},
  year={2019},
  organization={IEEE}
}

@inproceedings{abdelfattah2020best,
  title={Best of both worlds: Automl codesign of a cnn and its hardware accelerator},
  author={Abdelfattah, Mohamed S and Dudziak, {\L}ukasz and Chau, Thomas and Lee, Royson and Kim, Hyeji and Lane, Nicholas D},
  booktitle={2020 57th ACM/IEEE Design Automation Conference (DAC)},
  pages={1--6},
  year={2020},
  organization={IEEE}
}

@inproceedings{lu2019neural,
  title={On Neural Architecture Search for Resource-Constrained Hardware Platforms.},
  author={Lu, Qing and Jiang, Weiwen and Xu, Xiaowei and Shi, Yiyu and Hu, Jingtong},
  booktitle={International Conference on Computer-Aided Design},
  year={2019}
}

@article{jiang2020standing,
  title={Standing on the shoulders of giants: Hardware and neural architecture co-search with hot start},
  author={Jiang, Weiwen and Yang, Lei and Dasgupta, Sakyasingha and Hu, Jingtong and Shi, Yiyu},
  journal={IEEE Transactions on Computer-Aided Design of Integrated Circuits and Systems},
  volume={39},
  number={11},
  pages={4154--4165},
  year={2020},
  publisher={IEEE}
}

@inproceedings{zhang2020fast,
  title={Fast hardware-aware neural architecture search},
  author={Zhang, Li Lyna and Yang, Yuqing and Jiang, Yuhang and Zhu, Wenwu and Liu, Yunxin},
  booktitle={Proceedings of the IEEE/CVF Conference on Computer Vision and Pattern Recognition Workshops},
  pages={692--693},
  year={2020}
}

@inproceedings{jiang2020hardware,
  title={Hardware-aware transformable architecture search with efficient search space},
  author={Jiang, Yuhang and Wang, Xin and Zhu, Wenwu},
  booktitle={2020 IEEE International Conference on Multimedia and Expo (ICME)},
  pages={1--6},
  year={2020},
  organization={IEEE}
}

@inproceedings{chen2020you,
  title={You only search once: A fast automation framework for single-stage dnn/accelerator co-design},
  author={Chen, Weiwei and Wang, Ying and Yang, Shuang and Liu, Chen and Zhang, Lei},
  booktitle={2020 Design, Automation \& Test in Europe Conference \& Exhibition (DATE)},
  pages={1283--1286},
  year={2020},
  organization={IEEE}
}

@inproceedings{chen2021neural,
  title={Neural Architecture Search on ImageNet in Four GPU Hours: A Theoretically Inspired Perspective},
  author={Chen, Wuyang and Gong, Xinyu and Wang, Zhangyang},
  booktitle={International Conference on Learning Representations (ICLR)},
  year={2021}
}

@article{jiang2020device,
  title={Device-circuit-architecture co-exploration for computing-in-memory neural accelerators},
  author={Jiang, Weiwen and Lou, Qiuwen and Yan, Zheyu and Yang, Lei and Hu, Jingtong and Hu, Xiaobo Sharon and Shi, Yiyu},
  journal={IEEE Transactions on Computers},
  volume={70},
  number={4},
  pages={595--605},
  year={2020},
  publisher={IEEE}
}

@article{teich2012hardware,
  title={Hardware/software codesign: The past, the present, and predicting the future},
  author={Teich, J{\"u}rgen},
  journal={Proceedings of the IEEE},
  volume={100},
  number={Special Centennial Issue},
  pages={1411--1430},
  year={2012},
  publisher={IEEE}
}

@inproceedings{proxyless,
  title={ProxylessNAS: Direct Neural Architecture Search on Target Task and Hardware},
  author={Cai, Han and Zhu, Ligeng and Han, Song},
  booktitle={International Conference on Learning Representations},
  year         = {2019}
}

@inproceedings{dong2020bench,
  title={NAS-Bench-201: Extending the Scope of Reproducible Neural Architecture Search},
  author={Dong, Xuanyi and Yang, Yi},
  booktitle={International Conference on Learning Representations},
  year={2020}
}

@article{hsu2018monas,
  title={Monas: Multi-objective neural architecture search using reinforcement learning},
  author={Hsu, Chi-Hung and Chang, Shu-Huan and Liang, Jhao-Hong and Chou, Hsin-Ping and Liu, Chun-Hao and Chang, Shih-Chieh and Pan, Jia-Yu and Chen, Yu-Ting and Wei, Wei and Juan, Da-Cheng},
  journal={arXiv preprint arXiv:1806.10332},
  year={2018}
}

@article{sood2019neunets,
  title={Neunets: An automated synthesis engine for neural network design},
  author={Sood, Atin and Elder, Benjamin and Herta, Benjamin and Xue, Chao and Bekas, Costas and Malossi, A Cristiano I and Saha, Debashish and Scheidegger, Florian and Venkataraman, Ganesh and Thomas, Gegi and others},
  journal={arXiv preprint arXiv:1901.06261},
  year={2019}
}

@article{liashchynskyi2019grid,
  title={Grid search, random search, genetic algorithm: a big comparison for NAS},
  author={Liashchynskyi, Petro and Liashchynskyi, Pavlo},
  journal={arXiv preprint arXiv:1912.06059},
  year={2019}
}

@inproceedings{runge2018learning,
  title={Learning to Design {RNA}},
  author={Frederic Runge and Danny Stoll and Stefan Falkner and Frank Hutter},
  booktitle={International Conference on Learning Representations},
  year={2019}
}

@article{Ghosh-Dastidar2009Spiking,
title={Spiking Neural Networks},
author={S. Ghosh-Dastidar and H. Adeli},
journal={International journal of neural systems},
year={2009},
volume={19 4},
pages={ 295-308 },
doi={10.1142/S0129065709002002}
}

@article{nunes2022spiking,
  title={Spiking neural networks: A survey},
  author={Nunes, Joao D and Carvalho, Marcelo and Carneiro, Diogo and Cardoso, Jaime S},
  journal={IEEE Access},
  volume={10},
  pages={60738--60764},
  year={2022},
  publisher={IEEE}
}

@article{izhikevich2003simple,
  title={Simple model of spiking neurons},
  author={Izhikevich, Eugene M},
  journal={IEEE Transactions on neural networks},
  volume={14},
  number={6},
  pages={1569--1572},
  year={2003},
  publisher={IEEE}
}

@article{gao2007cortical,
  title={Cortical models onto CMOL and CMOS—architectures and performance/price},
  author={Gao, Changjian and Hammerstrom, Dan},
  journal={IEEE Transactions on Circuits and Systems I: Regular Papers},
  volume={54},
  number={11},
  pages={2502--2515},
  year={2007},
  publisher={IEEE}
}

@article{deng2020rethinking,
  title={Rethinking the performance comparison between SNNS and ANNS},
  author={Deng, Lei and Wu, Yujie and Hu, Xing and Liang, Ling and Ding, Yufei and Li, Guoqi and Zhao, Guangshe and Li, Peng and Xie, Yuan},
  journal={Neural networks},
  volume={121},
  pages={294--307},
  year={2020},
  publisher={Elsevier}
}

@article{izhikevich2003bursts,
  title={Bursts as a unit of neural information: selective communication via resonance},
  author={Izhikevich, Eugene M and Desai, Niraj S and Walcott, Elisabeth C and Hoppensteadt, Frank C},
  journal={Trends in neurosciences},
  volume={26},
  number={3},
  pages={161--167},
  year={2003},
  publisher={Elsevier}
}

@article{pfeiffer2018deep,
  title={Deep learning with spiking neurons: opportunities and challenges},
  author={Pfeiffer, Michael and Pfeil, Thomas},
  journal={Frontiers in neuroscience},
  volume={12},
  pages={409662},
  year={2018},
  publisher={Frontiers}
}

@inproceedings{valadez2017step,
  title={The step size impact on the computational cost of spiking neuron simulation},
  author={Valadez-God{\'\i}nez, Sergio and Sossa, Humberto and Santiago-Montero, Ra{\'u}l},
  booktitle={2017 Computing Conference},
  pages={722--728},
  year={2017},
  organization={IEEE}
}

@article{taherkhani2020review,
  title={A review of learning in biologically plausible spiking neural networks},
  author={Taherkhani, Aboozar and Belatreche, Ammar and Li, Yuhua and Cosma, Georgina and Maguire, Liam P and McGinnity, T Martin},
  journal={Neural Networks},
  volume={122},
  pages={253--272},
  year={2020},
  publisher={Elsevier}
}

@article{tavanaei2019deep,
  title={Deep learning in spiking neural networks},
  author={Tavanaei, Amirhossein and Ghodrati, Masoud and Kheradpisheh, Saeed Reza and Masquelier, Timoth{\'e}e and Maida, Anthony},
  journal={Neural networks},
  volume={111},
  pages={47--63},
  year={2019},
  publisher={Elsevier}
}

@article{wu2018spatio,
  title={Spatio-temporal backpropagation for training high-performance spiking neural networks},
  author={Wu, Yujie and Deng, Lei and Li, Guoqi and Zhu, Jun and Shi, Luping},
  journal={Frontiers in neuroscience},
  volume={12},
  pages={331},
  year={2018},
  publisher={Frontiers Media SA}
}

@article{neftci2019surrogate,
  title={Surrogate gradient learning in spiking neural networks: Bringing the power of gradient-based optimization to spiking neural networks},
  author={Neftci, Emre O and Mostafa, Hesham and Zenke, Friedemann},
  journal={IEEE Signal Processing Magazine},
  volume={36},
  number={6},
  pages={51--63},
  year={2019},
  publisher={IEEE}
}

@inproceedings{deng2023surrogate,
  title={Surrogate module learning: Reduce the gradient error accumulation in training spiking neural networks},
  author={Deng, Shikuang and Lin, Hao and Li, Yuhang and Gu, Shi},
  booktitle={International Conference on Machine Learning},
  pages={7645--7657},
  year={2023},
  organization={PMLR}
}

@inproceedings{schuman2020evolutionary,
  title={Evolutionary optimization for neuromorphic systems},
  author={Schuman, Catherine D and Mitchell, J Parker and Patton, Robert M and Potok, Thomas E and Plank, James S},
  booktitle={Proceedings of the 2020 Annual Neuro-Inspired Computational Elements Workshop},
  pages={1--9},
  year={2020}
}

@inproceedings{vigneron2020critical,
  title={A critical survey of STDP in Spiking Neural Networks for Pattern Recognition},
  author={Vigneron, Alex and Martinet, Jean},
  booktitle={2020 international joint conference on neural networks (ijcnn)},
  pages={1--9},
  year={2020},
  organization={IEEE}
}

@article{liu2021sstdp,
  title={SSTDP: Supervised spike timing dependent plasticity for efficient spiking neural network training},
  author={Liu, Fangxin and Zhao, Wenbo and Chen, Yongbiao and Wang, Zongwu and Yang, Tao and Jiang, Li},
  journal={Frontiers in Neuroscience},
  volume={15},
  pages={756876},
  year={2021},
  publisher={Frontiers Media SA}
}

@inproceedings{bohte2000spikeprop,
  title={SpikeProp: backpropagation for networks of spiking neurons.},
  author={Bohte, Sander M and Kok, Joost N and La Poutr{\'e}, Johannes A},
  booktitle={ESANN},
  volume={48},
  pages={419--424},
  year={2000},
  organization={Bruges}
}

@article{beyeler2013categorization,
  title={Categorization and decision-making in a neurobiologically plausible spiking network using a STDP-like learning rule},
  author={Beyeler, Michael and Dutt, Nikil D and Krichmar, Jeffrey L},
  journal={Neural Networks},
  volume={48},
  pages={109--124},
  year={2013},
  publisher={Elsevier}
}

@article{masquelier2007unsupervised,
  title={Unsupervised learning of visual features through spike timing dependent plasticity},
  author={Masquelier, Timoth{\'e}e and Thorpe, Simon J},
  journal={PLoS computational biology},
  volume={3},
  number={2},
  pages={e31},
  year={2007},
  publisher={Public Library of Science San Francisco, USA}
}

@inproceedings{tavanaei2016acquisition,
  title={Acquisition of visual features through probabilistic spike-timing-dependent plasticity},
  author={Tavanaei, Amirhossein and Masquelier, Timoth{\'e}e and Maida, Anthony S},
  booktitle={2016 International Joint Conference on Neural Networks (IJCNN)},
  pages={307--314},
  year={2016},
  organization={IEEE}
}

@article{kim2018deep,
  title={Deep neural networks with weighted spikes},
  author={Kim, Jaehyun and Kim, Heesu and Huh, Subin and Lee, Jinho and Choi, Kiyoung},
  journal={Neurocomputing},
  volume={311},
  pages={373--386},
  year={2018},
  publisher={Elsevier}
}

@inproceedings{wang2022signed,
  title={Signed Neuron with Memory: Towards Simple, Accurate and High-Efficient ANN-SNN Conversion.},
  author={Wang, Yuchen and Zhang, Malu and Chen, Yi and Qu, Hong},
  booktitle={IJCAI},
  pages={2501--2508},
  year={2022}
}

@inproceedings{deng2021optimal,
  title={Optimal Conversion of Conventional Artificial Neural Networks to Spiking Neural Networks},
  author={Deng, Shikuang and Gu, Shi},
  booktitle={International Conference on Learning Representations},
  year={2021}
}

@INPROCEEDINGS{ho2021conversion,
  author={Ho, Nguyen-Dong and Chang, Ik-Joon},
  booktitle={2021 58th ACM/IEEE Design Automation Conference (DAC)}, 
  title={TCL: an ANN-to-SNN Conversion with Trainable Clipping Layers}, 
  year={2021},
  volume={},
  number={},
  pages={793-798},
  doi={10.1109/DAC18074.2021.9586266}}

@article{rueckauer2017conversion,
  title={Conversion of continuous-valued deep networks to efficient event-driven networks for image classification},
  author={Rueckauer, Bodo and Lungu, Iulia-Alexandra and Hu, Yuhuang and Pfeiffer, Michael and Liu, Shih-Chii},
  journal={Frontiers in neuroscience},
  volume={11},
  pages={682},
  year={2017},
  publisher={Frontiers Media SA}
}

@inproceedings{sheik2017spike,
  title={Spike Based Information Processing in Spiking Neural Networks},
  author={Sheik, Sadique},
  booktitle={Proceedings of the 4th International Conference on Applications in Nonlinear Dynamics (ICAND 2016) 5},
  pages={177--188},
  year={2017},
  organization={Springer}
}

@article{almomani2019comparative,
  title={A comparative study on spiking neural network encoding schema: implemented with cloud computing},
  author={Almomani, Ammar and Alauthman, Mohammad and Alweshah, Mohammed and Dorgham, Osama and Albalas, Firas},
  journal={Cluster Computing},
  volume={22},
  pages={419--433},
  year={2019},
  publisher={Springer}
}

@inproceedings{park2020t2fsnn,
  title={T2FSNN: Deep spiking neural networks with time-to-first-spike coding},
  author={Park, Seongsik and Kim, Seijoon and Na, Byunggook and Yoon, Sungroh},
  booktitle={2020 57th ACM/IEEE Design Automation Conference (DAC)},
  pages={1--6},
  year={2020},
  organization={IEEE}
}

@article{guo2021neural,
  title={Neural coding in spiking neural networks: A comparative study for robust neuromorphic systems},
  author={Guo, Wenzhe and Fouda, Mohammed E and Eltawil, Ahmed M and Salama, Khaled Nabil},
  journal={Frontiers in Neuroscience},
  volume={15},
  pages={638474},
  year={2021},
  publisher={Frontiers Media SA}
}

@inproceedings{rueckauer2021temporal,
  title={Temporal pattern coding in deep spiking neural networks},
  author={Rueckauer, Bodo and Liu, Shih-Chii},
  booktitle={2021 International Joint Conference on Neural Networks (IJCNN)},
  pages={1--8},
  year={2021},
  organization={IEEE}
}

@article{bouvier2019spiking_survey,
  title={Spiking neural networks hardware implementations and challenges: A survey},
  author={Bouvier, Maxence and Valentian, Alexandre and Mesquida, Thomas and Rummens, Francois and Reyboz, Marina and Vianello, Elisa and Beigne, Edith},
  journal={ACM Journal on Emerging Technologies in Computing Systems (JETC)},
  volume={15},
  number={2},
  pages={1--35},
  year={2019},
  publisher={ACM New York, NY, USA}
}

@article{javanshir2022advancements,
  title={Advancements in algorithms and neuromorphic hardware for spiking neural networks},
  author={Javanshir, Amirhossein and Nguyen, Thanh Thi and Mahmud, MA Parvez and Kouzani, Abbas Z},
  journal={Neural Computation},
  volume={34},
  number={6},
  pages={1289--1328},
  year={2022},
  publisher={MIT Press One Rogers Street, Cambridge, MA 02142-1209, USA journals-info~…}
}

@article{truenorth,
  title={Truenorth: Design and tool flow of a 65 mw 1 million neuron programmable neurosynaptic chip},
  author={Akopyan, Filipp and Sawada, Jun and Cassidy, Andrew and Alvarez-Icaza, Rodrigo and Arthur, John and Merolla, Paul and Imam, Nabil and Nakamura, Yutaka and Datta, Pallab and Nam, Gi-Joon and others},
  journal={IEEE transactions on computer-aided design of integrated circuits and systems},
  volume={34},
  number={10},
  pages={1537--1557},
  year={2015},
  publisher={IEEE}
}

@article{davies2018loihi,
  title={Loihi: A neuromorphic manycore processor with on-chip learning},
  author={Davies, Mike and Srinivasa, Narayan and Lin, Tsung-Han and Chinya, Gautham and Cao, Yongqiang and Choday, Sri Harsha and Dimou, Georgios and Joshi, Prasad and Imam, Nabil and Jain, Shweta and others},
  journal={Ieee Micro},
  volume={38},
  number={1},
  pages={82--99},
  year={2018},
  publisher={IEEE}
}

@article{abderrahmane2020design,
  title={Design space exploration of hardware spiking neurons for embedded artificial intelligence},
  author={Abderrahmane, Nassim and Lemaire, Edgar and Miramond, Beno{\^\i}t},
  journal={Neural Networks},
  volume={121},
  pages={366--386},
  year={2020},
  publisher={Elsevier}
}

@article{painkras2013spinnaker,
  title={SpiNNaker: A 1-W 18-core system-on-chip for massively-parallel neural network simulation},
  author={Painkras, Eustace and Plana, Luis A and Garside, Jim and Temple, Steve and Galluppi, Francesco and Patterson, Cameron and Lester, David R and Brown, Andrew D and Furber, Steve B},
  journal={IEEE Journal of Solid-State Circuits},
  volume={48},
  number={8},
  pages={1943--1953},
  year={2013},
  publisher={IEEE}
}

@article{rathi2023exploring,
  title={Exploring neuromorphic computing based on spiking neural networks: Algorithms to hardware},
  author={Rathi, Nitin and Chakraborty, Indranil and Kosta, Adarsh and Sengupta, Abhronil and Ankit, Aayush and Panda, Priyadarshini and Roy, Kaushik},
  journal={ACM Computing Surveys},
  volume={55},
  number={12},
  pages={1--49},
  year={2023},
  publisher={ACM New York, NY}
}

@article{ma2017darwin,
  title={Darwin: A neuromorphic hardware co-processor based on spiking neural networks},
  author={Ma, De and Shen, Juncheng and Gu, Zonghua and Zhang, Ming and Zhu, Xiaolei and Xu, Xiaoqiang and Xu, Qi and Shen, Yangjing and Pan, Gang},
  journal={Journal of systems architecture},
  volume={77},
  pages={43--51},
  year={2017},
  publisher={Elsevier}
}

@article{li2021fast,
  title={A fast and energy-efficient SNN processor with adaptive clock/event-driven computation scheme and online learning},
  author={Li, Sixu and Zhang, Zhaomin and Mao, Ruixin and Xiao, Jianbiao and Chang, Liang and Zhou, Jun},
  journal={IEEE Transactions on Circuits and Systems I: Regular Papers},
  volume={68},
  number={4},
  pages={1543--1552},
  year={2021},
  publisher={IEEE}
}

@article{hsieh2012vlsi,
  title={VLSI implementation of a bio-inspired olfactory spiking neural network},
  author={Hsieh, Hung-Yi and Tang, Kea-Tiong},
  journal={IEEE transactions on neural networks and learning systems},
  volume={23},
  number={7},
  pages={1065--1073},
  year={2012},
  publisher={IEEE}
}

@inproceedings{schemmel2010wafer,
  title={A wafer-scale neuromorphic hardware system for large-scale neural modeling},
  author={Schemmel, Johannes and Br{\"u}derle, Daniel and Gr{\"u}bl, Andreas and Hock, Matthias and Meier, Karlheinz and Millner, Sebastian},
  booktitle={2010 IEEE International Symposium on Circuits and Systems (ISCAS)},
  pages={1947--1950},
  year={2010},
  organization={IEEE}
}

@article{qu2020high,
  title={High performance simulation of spiking neural network on gpgpus},
  author={Qu, Peng and Zhang, Youhui and Fei, Xiang and Zheng, Weimin},
  journal={IEEE Transactions on Parallel and Distributed Systems},
  volume={31},
  number={11},
  pages={2510--2523},
  year={2020},
  publisher={IEEE}
}

@article{han2020hardware,
  title={Hardware implementation of spiking neural networks on FPGA},
  author={Han, Jianhui and Li, Zhaolin and Zheng, Weimin and Zhang, Youhui},
  journal={Tsinghua Science and Technology},
  volume={25},
  number={4},
  pages={479--486},
  year={2020},
  publisher={TUP}
}

@article{pani2017fpga,
  title={An FPGA platform for real-time simulation of spiking neuronal networks},
  author={Pani, Danilo and Meloni, Paolo and Tuveri, Giuseppe and Palumbo, Francesca and Massobrio, Paolo and Raffo, Luigi},
  journal={Frontiers in neuroscience},
  volume={11},
  pages={90},
  year={2017},
  publisher={Frontiers Media SA}
}

@article{liu2022fpga,
  title={FPGA-NHAP: A general FPGA-based neuromorphic hardware acceleration platform with high speed and low power},
  author={Liu, Yijun and Chen, Yuehai and Ye, Wujian and Gui, Yu},
  journal={IEEE Transactions on Circuits and Systems I: Regular Papers},
  volume={69},
  number={6},
  pages={2553--2566},
  year={2022},
  publisher={IEEE}
}

@article{ju2020fpga,
  title={An FPGA implementation of deep spiking neural networks for low-power and fast classification},
  author={Ju, Xiping and Fang, Biao and Yan, Rui and Xu, Xiaoliang and Tang, Huajin},
  journal={Neural computation},
  volume={32},
  number={1},
  pages={182--204},
  year={2020},
  publisher={MIT Press One Rogers Street, Cambridge, MA 02142-1209, USA journals-info~…}
}

@inproceedings{aliyev2024fine,
  title={Fine-tuning surrogate gradient learning for optimal hardware performance in spiking neural networks},
  author={Aliyev, Ilkin and Adegbija, Tosiron},
  booktitle={2024 Design, Automation \& Test in Europe Conference \& Exhibition (DATE)},
  pages={1--2},
  year={2024},
  organization={IEEE}
}

@inproceedings{aliyev2024_sparsityOverview,
  title={Sparsity-Aware Hardware-Software Co-Design of Spiking Neural Networks: An Overview},
  author={Aliyev, Ilkin and Svoboda, Kama and Adegbija, Tosiron and Fellous, Jean-Marc},
  booktitle={2024 IEEE 17th International Symposium on Embedded Multicore/Many-core Systems-on-Chip (MCSoC)},
  pages={413--420},
  year={2024},
  organization={IEEE}
}

@article{aliyev2025exploring,
  title={Exploring the Sparsity-Quantization Interplay on a Novel Hybrid SNN Event-Driven Architecture},
  author={Aliyev, Ilkin and Lopez, Jesus and Adegbija, Tosiron},
  booktitle={2025 Design, Automation \& Test in Europe Conference \& Exhibition (DATE)},
  year={2025},
  organization={IEEE}
}

@article{yavuz2016genn,
  title={GeNN: a code generation framework for accelerated brain simulations},
  author={Yavuz, Esin and Turner, James and Nowotny, Thomas},
  journal={Scientific reports},
  volume={6},
  number={1},
  pages={18854},
  year={2016},
  publisher={Nature Publishing Group UK London}
}

@inproceedings{beyeler2015carlsim,
  title={CARLsim 3: A user-friendly and highly optimized library for the creation of neurobiologically detailed spiking neural networks},
  author={Beyeler, Michael and Carlson, Kristofor D and Chou, Ting-Shuo and Dutt, Nikil and Krichmar, Jeffrey L},
  booktitle={2015 International Joint Conference on Neural Networks (IJCNN)},
  pages={1--8},
  year={2015},
  organization={IEEE}
}

@article{rathi2021diet,
  title={Diet-snn: A low-latency spiking neural network with direct input encoding and leakage and threshold optimization},
  author={Rathi, Nitin and Roy, Kaushik},
  journal={IEEE Transactions on Neural Networks and Learning Systems},
  year={2021},
  publisher={IEEE}
}

@article{cramer2020heidelberg,
  title={The heidelberg spiking data sets for the systematic evaluation of spiking neural networks},
  author={Cramer, Benjamin and Stradmann, Yannik and Schemmel, Johannes and Zenke, Friedemann},
  journal={IEEE Transactions on Neural Networks and Learning Systems},
  volume={33},
  number={7},
  pages={2744--2757},
  year={2020},
  publisher={IEEE}
}

@mastersthesis{masters,
  author={Nijenhuis, J.},
  title={Using NAS to improve accuracy of SNNs},
  school = {Eindhoven University of Technology},
  type = {M.S. thesis},
  address = {Eindhoven, Netherlands},
  year = {2021}
}

@article{che2023auto,
  title={Auto-spikformer: Spikformer architecture search},
  author={Che, Kaiwei and Zhou, Zhaokun and Niu, Jun and Ma, Zhengyu and Fang, Wei and Chen, Yanqi and Shen, Shuaijie and Yuan, Li and Tian, Yonghong},
  journal={Frontiers in Neuroscience},
  volume={18},
  pages={1372257},
  year={2024},
  publisher={Frontiers Media SA}
}

@ARTICLE{pan2023multi,
  author={Pan, Wenxuan and Zhao, Feifei and Shen, Guobin and Han, Bing and Zeng, Yi},
  journal={IEEE Transactions on Evolutionary Computation}, 
  title={Brain-Inspired Multiscale Evolutionary Neural Architecture Search for Deep Spiking Neural Networks}, 
  year={2025},
  volume={29},
  number={5},
  pages={2258-2270},
  doi={10.1109/TEVC.2024.3507812}
}

@inproceedings{anwar2021evolving,
  title={Evolving Spiking Circuit Motifs Using Weight Agnostic Neural Networks},
  author={Anwar, Abrar},
  booktitle={Proceedings of the AAAI Conference on Artificial Intelligence},
  volume={35},
  number={18},
  pages={15956--15957},
  year={2021}
}

@article{pan2024brain,
  title={Brain-inspired Evolutionary Architectures for Spiking Neural Networks},
  author={Pan, Wenxuan and Zhao, Feifei and Zhao, Zhuoya and Zeng, Yi},
  journal={IEEE Transactions on Artificial Intelligence},
  year={2024},
  publisher={IEEE}
}

@article{putra2024methodology,
  title={Spikernel: A kernel size exploration methodology for improving accuracy of the embedded spiking neural network systems},
  author={Putra, Rachmad Vidya Wicaksana and Shafique, Muhammad},
  journal={IEEE Embedded Systems Letters},
  year={2024},
  publisher={IEEE}
}

@article{putra2024neuronas,
  title={NeuroNAS: Enhancing Efficiency of Neuromorphic In-Memory Computing for Intelligent Mobile Agents through Hardware-Aware Spiking Neural Architecture Search},
  author={Putra, Rachmad Vidya Wicaksana and Shafique, Muhammad},
  journal={arXiv preprint arXiv:2407.00641},
  year={2024}
}

@article{putra2025spikenas,
  title={SpikeNAS: A Fast Memory-Aware Neural Architecture Search Framework for Spiking Neural Network-based Embedded AI Systems},
  author={Putra, Rachmad Vidya Wicaksana and Shafique, Muhammad},
  journal={IEEE Transactions on Artificial Intelligence},
  year={2025},
  publisher={IEEE}
}

@inproceedings{xie2023efficient,
  title={Efficient Spiking Neural Architecture Search with Mixed Neuron Models and Variable Thresholds},
  author={Xie, Zaipeng and Liu, Ziang and Chen, Peng and Zhang, Jianan},
  booktitle={International Conference on Neural Information Processing},
  pages={466--481},
  year={2023},
  organization={Springer}
}

@article{che2022differentiable,
  title={Differentiable hierarchical and surrogate gradient search for spiking neural networks},
  author={Che, Kaiwei and Leng, Luziwei and Zhang, Kaixuan and Zhang, Jianguo and Meng, Qinghu and Cheng, Jie and Guo, Qinghai and Liao, Jianxing},
  journal={Advances in Neural Information Processing Systems},
  volume={35},
  pages={24975--24990},
  year={2022}
}

@inproceedings{liu2024unleashing,
  title={Unleashing Energy-Efficiency: Neural Architecture Search without Training for Spiking Neural Networks on Loihi Chip},
  author={Liu, Shiya and Yi, Yang},
  booktitle={2024 25th International Symposium on Quality Electronic Design (ISQED)},
  pages={1--7},
  year={2024},
  organization={IEEE}
}

@inproceedings{wang2024autost,
  title={AutoST: Training-free Neural Architecture Search for Spiking Transformers},
  author={Wang, Ziqing and Zhao, Qidong and Cui, Jinku and Liu, Xu and Xu, Dongkuan},
  booktitle={ICASSP 2024-2024 IEEE International Conference on Acoustics, Speech and Signal Processing (ICASSP)},
  pages={3455--3459},
  year={2024},
  organization={IEEE}
}

@inproceedings{ajay2024despine,
  title={DESPINE: NAS generated Deep Evolutionary Adaptive Spiking Network for Low Power Edge Computing Applications},
  author={Ajay, BS and Rao, Madhav and others},
  booktitle={2024 25th International Symposium on Quality Electronic Design (ISQED)},
  pages={1--8},
  year={2024},
  organization={IEEE}
}

@article{li2024spiking,
  title={Spiking Spatio-Temporal Neural Architecture Search for EEG-Based Emotion Recognition},
  author={Li, Wei and Zhu, Zhihao and Shao, Shitong and Lu, Yao and Song, Aiguo},
  journal={IEEE Transactions on Instrumentation and Measurement},
  year={2024},
  publisher={IEEE}
}

@article{padovano2024spikeexplorer,
  title={SpikeExplorer: Hardware-Oriented Design Space Exploration for Spiking Neural Networks on FPGA},
  author={Padovano, Dario and Carpegna, Alessio and Savino, Alessandro and Di Carlo, Stefano},
  journal={Electronics},
  volume={13},
  number={9},
  pages={1744},
  year={2024},
  publisher={MDPI}
}

@inproceedings{kim2022neural,
  title={Neural architecture search for spiking neural networks},
  author={Kim, Youngeun and Li, Yuhang and Park, Hyoungseob and Venkatesha, Yeshwanth and Panda, Priyadarshini},
  booktitle={European Conference on Computer Vision},
  pages={36--56},
  year={2022},
  organization={Springer}
}

@article{shen2023brain,
  title={Brain-inspired neural circuit evolution for spiking neural networks},
  author={Shen, Guobin and Zhao, Dongcheng and Dong, Yiting and Zeng, Yi},
  journal={Proceedings of the National Academy of Sciences},
  volume={120},
  number={39},
  pages={e2218173120},
  year={2023},
  publisher={National Academy of Sciences}
}

@article{shen2024evolutionary,
  title={Evolutionary spiking neural networks: a survey},
  author={Shen, Shuaijie and Zhang, Rui and Wang, Chao and Huang, Renzhuo and Tuerhong, Aiersi and Guo, Qinghai and Lu, Zhichao and Zhang, Jianguo and Leng, Luziwei},
  journal={Journal of Membrane Computing},
  pages={1--12},
  year={2024},
  publisher={Springer}
}

@inproceedings{liu2024lite,
  title={LitE-SNN: designing lightweight and efficient spiking neural network through spatial-temporal compressive network search and joint optimization},
  author={Liu, Qianhui and Yan, Jiaqi and Zhang, Malu and Pan, Gang and Li, Haizhou},
  booktitle={Proceedings of the Thirty-Third International Joint Conference on Artificial Intelligence},
  pages={3097--3105},
  year={2024}
}

@article{yan2024efficient,
  title={Efficient spiking neural network design via neural architecture search},
  author={Yan, Jiaqi and Liu, Qianhui and Zhang, Malu and Feng, Lang and Ma, De and Li, Haizhou and Pan, Gang},
  journal={Neural Networks},
  pages={106172},
  year={2024},
  publisher={Elsevier}
}

@article{che2024spatial,
  title={Spatial-Temporal Search for Spiking Neural Networks},
  author={Che, Kaiwei and Zhou, Zhaokun and Yuan, Li and Zhang, Jianguo and Tian, Yonghong and Leng, Luziwei},
  journal={arXiv preprint arXiv:2410.18580},
  year={2024}
}

@article{zhang2024ancoef,
  title={ANCoEF: Asynchronous Neuromorphic Algorithm/Hardware Co-Exploration Framework with a Fully Asynchronous Simulator},
  author={Zhang, Jian and Zhang, Xiang and Huang, Jingchen and Zhang, Jilin and Chen, Hong},
  journal={arXiv preprint arXiv:2411.06059},
  year={2024}
}

@INPROCEEDINGS{gaq,
  author={Nguyen, Duy-Anh and Tran, Xuan-Tu and Iacopi, Francesca},
  booktitle={2022 International Conference on IC Design and Technology (ICICDT)}, 
  title={GAQ-SNN: A Genetic Algorithm based Quantization Framework for Deep Spiking Neural Networks}, 
  year={2022},
  volume={},
  number={},
  pages={93-96},
  doi={10.1109/ICICDT56182.2022.9933070}}

@inproceedings{na2022autosnn,
  title={Autosnn: Towards energy-efficient spiking neural networks},
  author={Na, Byunggook and Mok, Jisoo and Park, Seongsik and Lee, Dongjin and Choe, Hyeokjun and Yoon, Sungroh},
  booktitle={International Conference on Machine Learning},
  pages={16253--16269},
  year={2022},
  organization={PMLR}
}

@article{mnist,
  title={Gradient-based learning applied to document recognition},
  author={LeCun, Yann and Bottou, L{\'e}on and Bengio, Yoshua and Haffner, Patrick},
  journal={Proceedings of the IEEE},
  volume={86},
  number={11},
  pages={2278--2324},
  year={1998},
  publisher={Ieee}
}

@article{song2025evolutionary,
    title={Evolutionary multi-objective spiking neural architecture search for image classification},
    author={Song, Xiaotian and Lv, Zeqiong and Fan, Jiaohao and Deng, Xiong and Lv, Jiancheng and Liu, Jiyuan and Sun, Yanan},
    journal={IEEE Transactions on Evolutionary Computation},
    year={2025},
    publisher={IEEE}
}

@inproceedings{saghand2025monas,
    title={MONAS-ESNN: Multi-Objective Neural Architecture Search for Efficient Spiking Neural Networks},
    author={Saghand, Esmat Ghasemi and Lai-Yuen, Susana K},
    booktitle={2025 IEEE/CVF Winter Conference on Applications of Computer Vision (WACV)},
    pages={2963--2972},
    year={2025},
    organization={IEEE}
}

@inproceedings{abdennadher2025lightsnn,
  title={LightSNN: Lightweight Architecture Search for Sparse and Accurate Spiking Neural Networks},
  author={Abdennadher, Yesmine and Perint, Giovanni and Mazzieri, Riccardo and Pegoraro, Jacopo and Rossi, Michele},
  booktitle={2025 International Conference on Advanced Machine Learning and Data Science (AMLDS)},
  pages={84--89},
  year={2025},
  organization={IEEE}
}

@article{sun2025spikenas,
    title={SpikeNAS-Bench: Benchmarking NAS Algorithms for Spiking Neural Network Architecture},
    author={Sun, Genchen and Liu, Zhengkun and Gan, Lin and Su, Hang and Li, Ting and Zhao, Wenfeng and Sun, Biao},
    journal={IEEE Transactions on Artificial Intelligence},
    year={2025},
    publisher={IEEE}
}

@article{wang2025predictor,
  title={Predictor-assisted Evolutionary Neural Architecture Search for Spiking Neural Networks},
  author={Wang, Yuting and Xue, Yu and Ding, Wei and Tan, Yiyu and Chen, Peng and Wahib, Mohamed},
  journal={Neurocomputing},
  pages={131244},
  year={2025},
  publisher={Elsevier}
}

@article{liu2025efficient,
  title={Efficient Eye-based Emotion Recognition via Neural Architecture Search of Time-to-First-Spike-Coded Spiking Neural Networks},
  author={Liu, Qianhui and Yang, Jing and Yu, Miao and Carlson, Trevor E and Pan, Gang and Li, Haizhou and Chen, Zhumin},
  journal={arXiv preprint arXiv:2512.02459},
  year={2025}
}

@article{liu2025auto,
  title={Auto Deep Spiking Neural Network Design Based on an Evolutionary Membrane Algorithm},
  author={Liu, Chuang and Wang, Haojie},
  journal={Biomimetics},
  volume={10},
  number={8},
  pages={514},
  year={2025},
  publisher={MDPI}
}

@article{man2024differentiable,
  title={Differentiable architecture search with multi-dimensional attention for spiking neural networks},
  author={Man, Yilei and Xie, Linhai and Qiao, Shushan and Zhou, Yumei and Shang, Delong},
  journal={Neurocomputing},
  volume={601},
  pages={128181},
  year={2024},
  publisher={Elsevier}
}

@article{putra2024spikernel,
  title={Spikernel: A kernel size exploration methodology for improving accuracy of the embedded spiking neural network systems},
  author={Putra, Rachmad Vidya Wicaksana and Shafique, Muhammad},
  journal={IEEE Embedded Systems Letters},
  year={2024},
  publisher={IEEE}
}

@article{zhang2025spike,
  title={Spike-Driven Lightweight Large Language Model With Evolutionary Computation},
  author={Zhang, Malu and Wei, Wenjie and Zhou, Zijian and Liu, Wanlong and Zhang, Jie and Belatreche, Ammar and Yang, Yang},
  journal={IEEE Transactions on Evolutionary Computation},
  year={2025},
  publisher={IEEE}
}

@inproceedings{yanguas2022automl,
  title={AutoML for neuromorphic computing and application-driven co-design: asynchronous, massively parallel optimization of spiking architectures},
  author={Yanguas-Gil, Angel and Madireddy, Sandeep},
  booktitle={2022 IEEE International Conference on Rebooting Computing (ICRC)},
  pages={24--29},
  year={2022},
  organization={IEEE}
}

@article{dengproper,
  title={Proper Backward Connection Placement Boosts Spiking Neural Networks},
  author={Deng, Shikuang and Li, Wei and Wu, Yuhang and Zheng, Peixiang and Xin, Du Zong and Gu, Shi},
year={2023}
}

@article{zhou2020surrogate,
  title={Surrogate-assisted evolutionary search of spiking neural architectures in liquid state machines},
  author={Zhou, Yan and Jin, Yaochu and Ding, Jinliang},
  journal={Neurocomputing},
  volume={406},
  pages={12--23},
  year={2020},
  publisher={Elsevier}
}

@inproceedings{shen2023esl,
  title={Esl-snns: An evolutionary structure learning strategy for spiking neural networks},
  author={Shen, Jiangrong and Xu, Qi and Liu, Jian K and Wang, Yueming and Pan, Gang and Tang, Huajin},
  booktitle={Proceedings of the AAAI Conference on Artificial Intelligence},
  volume={37},
  number={1},
  pages={86--93},
  year={2023}
}

@article{zhangspatio,
  title={Spatio-Temporal Dependency-Aware Neuron Optimization for Spiking Neural Networks},
  author={Zhang, Desong and Hu, Jia and Min, Geyong},
year={2025}
}

@article{xie2025spear,
  title={SPEAR: Structured Pruning for Spiking Neural Networks via Synaptic Operation Estimation and Reinforcement Learning},
  author={Xie, Hui and Liu, Yuhe and Yang, Shaoqi and Guo, Jinyang and Guo, Yufei and Ma, Yuqing and Chen, Jiaxin and Liu, Jiaheng and Liu, Xianglong},
  journal={arXiv preprint arXiv:2507.02945},
  year={2025}
}

@inproceedings{branquinho2023spenser,
  title={SPENSER: Towards a NeuroEvolutionary approach for convolutional spiking neural networks},
  author={Branquinho, Henrique and Louren{\c{c}}o, Nuno and Costa, Ernesto},
  booktitle={Proceedings of the Companion Conference on Genetic and Evolutionary Computation},
  pages={2115--2122},
  year={2023}
}

@article{deng2024spiking,
  title={Spiking Token Mixer: An event-driven friendly Former structure for spiking neural networks},
  author={Deng, Shikuang and Wu, Yuhang and Du, Kangrui and Gu, Shi},
  journal={Advances in Neural Information Processing Systems},
  volume={37},
  pages={128825--128846},
  year={2024}
}

@inproceedings{benmeziane2023skip,
  title={Skip connections in spiking neural networks: an analysis of their effect on network training},
  author={Benmeziane, Hadjer and Ounnoughene, Amine Ziad and Hamzaoui, Imane and Bouhadjar, Younes},
  booktitle={2023 IEEE International Parallel and Distributed Processing Symposium Workshops (IPDPSW)},
  pages={790--794},
  year={2023},
  organization={IEEE}
}

@article{xu2025spikex,
  title={SpikeX: Exploring Accelerator Architecture and Network-Hardware Co-Optimization for Sparse Spiking Neural Networks},
  author={Xu, Boxun and Boone, Richard and Li, Peng},
  journal={arXiv preprint arXiv:2505.12292},
  year={2025}
}

@inproceedings{vasilache2025evolving,
  title={Evolving Spatially Embedded Recurrent Spiking Neural Networks for Control Tasks},
  author={Vasilache, Alexandru and Scholz, Jona and Sandamirskaya, Yulia and Becker, J{\"u}rgen},
  booktitle={International Conference on Artificial Neural Networks},
  pages={66--77},
  year={2025},
  organization={Springer}
}

@misc{dimovska2019multi,
  title={Multi-objective optimization for size and resilience of spiking neural networks. In 2019 IEEE 10th Annual Ubiquitous Computing, Electronics \& Mobile Communication Conference (UEMCON)},
  author={Dimovska, Mihaela and Johnston, Travis and Schuman, Catherine D and Mitchell, J Parker and Potok, Thomas E},
  year={2019},
  publisher={IEEE}
}

@inproceedings{schuman2016evolutionary,
  title={An evolutionary optimization framework for neural networks and neuromorphic architectures},
  author={Schuman, Catherine D and Plank, James S and Disney, Adam and Reynolds, John},
  booktitle={2016 International Joint Conference on Neural Networks (IJCNN)},
  pages={145--154},
  year={2016},
  organization={IEEE}
}

@inproceedings{wang2025asnpc,
  title={ASNPC: An Automated Generation Framework for SNN and Neuromorphic Processor Co-Design},
  author={Wang, Xiangyu and Li, Yuan and Yang, Zhijie and Xiao, Chao and Xiao, Xun and Chen, Renzhi and Xu, Weixia and Wang, Lei},
  booktitle={2025 Design, Automation \& Test in Europe Conference (DATE)},
  pages={1--7},
  year={2025},
  organization={IEEE}
}

@article{chen2024based,
  title={A NAS-Based TinyML for Secure Authentication Detection on SAGVN-Enabled Consumer Edge Devices},
  author={Chen, Xingchi and Zhang, Shuanglong and Li, Qing and Zhu, Fa and Feng, Ansong and Nkenyereye, Lewis and Rani, Shalli},
  journal={IEEE Transactions on Consumer Electronics},
  year={2024},
  publisher={IEEE}
}

@phdthesis{buettner2021case,
  title={A Case Study in Practical Neuromorphic Computing: Heartbeat Classification on the Loihi Neuromorphic Processor},
  author={Buettner, Kyle},
  year={2021},
  school={University of Pittsburgh}
}

@inproceedings{balaji2020compiling,
  title={Compiling spiking neural networks to mitigate neuromorphic hardware constraints},
  author={Balaji, Adarsha and Das, Anup},
  booktitle={2020 11th International Green and Sustainable Computing Workshops (IGSC)},
  pages={1--3},
  year={2020},
  organization={IEEE}
}

@inproceedings{seekings2024towards,
  title={Towards efficient deployment of hybrid snns on neuromorphic and edge AI hardware},
  author={Seekings, James and Chandarana, Peyton and Ardakani, Mahsa and Mohammadi, MohammadReza and Zand, Ramtin},
  booktitle={2024 International Conference on Neuromorphic Systems (ICONS)},
  pages={71--77},
  year={2024},
  organization={IEEE}
}

@article{towers2022long,
  title={Long-term reproducibility for neural architecture search},
  author={Towers, David and Forshaw, Matthew and Atapour-Abarghouei, Amir and McGough, Andrew Stephen},
  journal={arXiv preprint arXiv:2207.04821},
  year={2022}
}
}

\end{document}